\documentclass[journal]{IEEEtran}

\usepackage{cite}

\usepackage{bbm}
\usepackage{xcolor}
\usepackage{caption}
\usepackage{subcaption}
\usepackage{url}
\usepackage{amssymb}
\usepackage{bm}
\usepackage{amsfonts}
\usepackage{graphicx}
\newcommand{\gtlt}{{\left. \begin{array}{ll} H_0  \\ >   \\ <   \\ H_1  \\ \end{array} \right. }}

\newtheorem{proposition}{Proposition}
\newtheorem{observation}{Observation}
\usepackage[utf8]{inputenc} 
\usepackage[T1]{fontenc}
\usepackage{url}
\usepackage{ifthen}
\usepackage[cmex10]{amsmath}

\begin{document}
\title{Generalized Likelihood Ratio Test for Adversarially Robust Hypothesis Testing}

\author{Bhagyashree~Puranik,~\IEEEmembership{Student Member,~IEEE,}
	Upamanyu~Madhow,~\IEEEmembership{Fellow,~IEEE,}
	and~Ramtin~Pedarsani,~\IEEEmembership{Senior Member,~IEEE}
	\thanks{This manuscript was presented in part at the 2021 IEEE International Conference on Acoustics, Speech and Signal Processing (ICASSP)~\cite{puranik}.}
	\thanks{B. Puranik, U. Madhow and R. Pedarsani are with the Department
		of Electrical and Computer Engineering, University of California Santa Barbara, Santa Barbara, CA, 93106 USA (e-mail: bpuranik@ucsb.edu, madhow@ucsb.edu, ramtin@ucsb.edu).}}

\interdisplaylinepenalty=2500 

\maketitle

\begin{abstract}
Machine learning models are known to be susceptible to adversarial attacks which can cause misclassification by introducing small but well designed perturbations. In this paper, we consider a classical hypothesis testing problem in order to develop fundamental insight into defending against such adversarial perturbations. We interpret an adversarial perturbation
as a nuisance parameter, and propose a defense based on applying the generalized likelihood ratio test (GLRT) to the resulting composite hypothesis testing problem, jointly estimating the class of interest and the adversarial perturbation.  While the GLRT approach is applicable to general multi-class hypothesis testing, we first evaluate it for binary hypothesis testing in white Gaussian noise under $\ell_{\infty}$ norm-bounded adversarial perturbations, for which a known minimax defense optimizing for the worst-case attack provides a benchmark. We derive the worst-case attack for the GLRT defense, and show that its asymptotic performance (as the dimension of the data increases) approaches that of the minimax defense. For non-asymptotic regimes, we show via simulations that the GLRT defense is competitive with the minimax approach under the worst-case attack, while yielding a better robustness-accuracy tradeoff under weaker attacks. We also illustrate the GLRT approach for a multi-class hypothesis testing problem, for which a minimax strategy is not known, evaluating its performance under both noise-agnostic and noise-aware adversarial settings, by providing a method to find optimal noise-aware attacks, and heuristics to find noise-agnostic attacks that are close to optimal in the high SNR regime.

\end{abstract}


\IEEEpeerreviewmaketitle

\section{Introduction}
\label{sec:intro}

\IEEEPARstart{W}{hile} discussion of security in machine learning predates deep learning~\cite{barreno}, it becomes critical
to address these concerns in view of the widespread adoption of deep neural networks in safety- and security-critical applications such as facial recognition for surveillance, autonomous driving and virtual assistants.  In particular, it is known that deep neural networks are vulnerable to {\it adversarial attacks:} an adversary is often able to add small perturbations to data in an intelligent way to cause misclassification with high confidence ~\cite{szegedy,rolli}. Studies have shown that adversarial examples exist even in real-world physical systems. For example, an adversarial attack can manipulate traffic signs to fool autonomous vehicles~\cite{kurakin_17} or tamper with speech recognition systems~\cite{carlini_16,ravanelli}. In applications that demand robustness, such adversarial attacks are fundamental threats, which motivates a rapidly growing body of research on both attacks and defenses.  Some defenses are certifiably robust~\cite{wong,liang1}, while others are empirical~\cite{madry_iclr2018, goodfellow_2015}.  Many suggested defenses have been broken by subsequent attacks~\cite{tramer_20, carlini2017,carlini2018}. The present state of the art defenses ~\cite{madry_iclr2018, zhang_19, carmon_19} are purely empirical, relying on {\it adversarial training,} wherein adversarial perturbations are applied while training the neural network.  However, we do not yet have robustness guarantees or structural insights for such adversarially trained networks. Thus, existing defenses may be prone to new attacks that are conceived in future, possibly taking advantage of the availability of increased computational power~\cite{bubeck_icml19}.  It is essential, therefore, to develop at least a statistical understanding of the robustness that can be provided by a classifier.

In this paper, we take a step back from deep neural networks, and attempt to  develop fundamental insight into the impact of adversarial attacks on classification performance
in the framework of classical hypothesis testing.  Specifically, we investigate adversarial classification in the setting of composite hypothesis testing, in which the class-conditional distributions of the data are known, and the adversarial perturbation is treated as a nuisance parameter. We adopt a Generalized Likelihood Ratio Test (GLRT) formulation for defense against adversarial attacks, in which we jointly estimate the desired class and the action of the adversary. 

\begin{figure}[b!]
	\centering
	\begin{subfigure}[t]{0.45\columnwidth}
		\centering
		\includegraphics[width=\columnwidth]{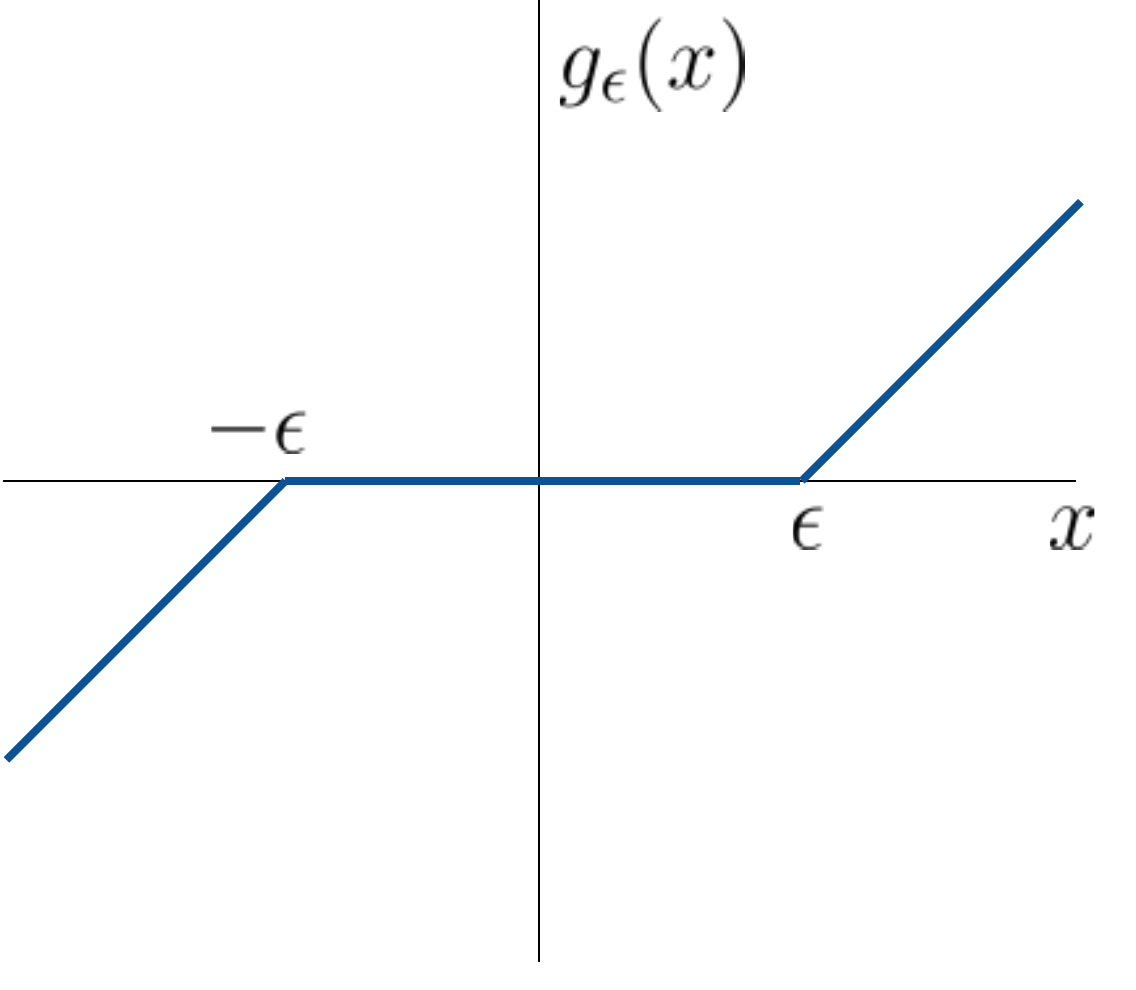}
		\caption{}
		\label{fig:g_eps}
	\end{subfigure}
	\begin{subfigure}[t]{0.45\columnwidth}
		\centering
		\includegraphics[width=\columnwidth]{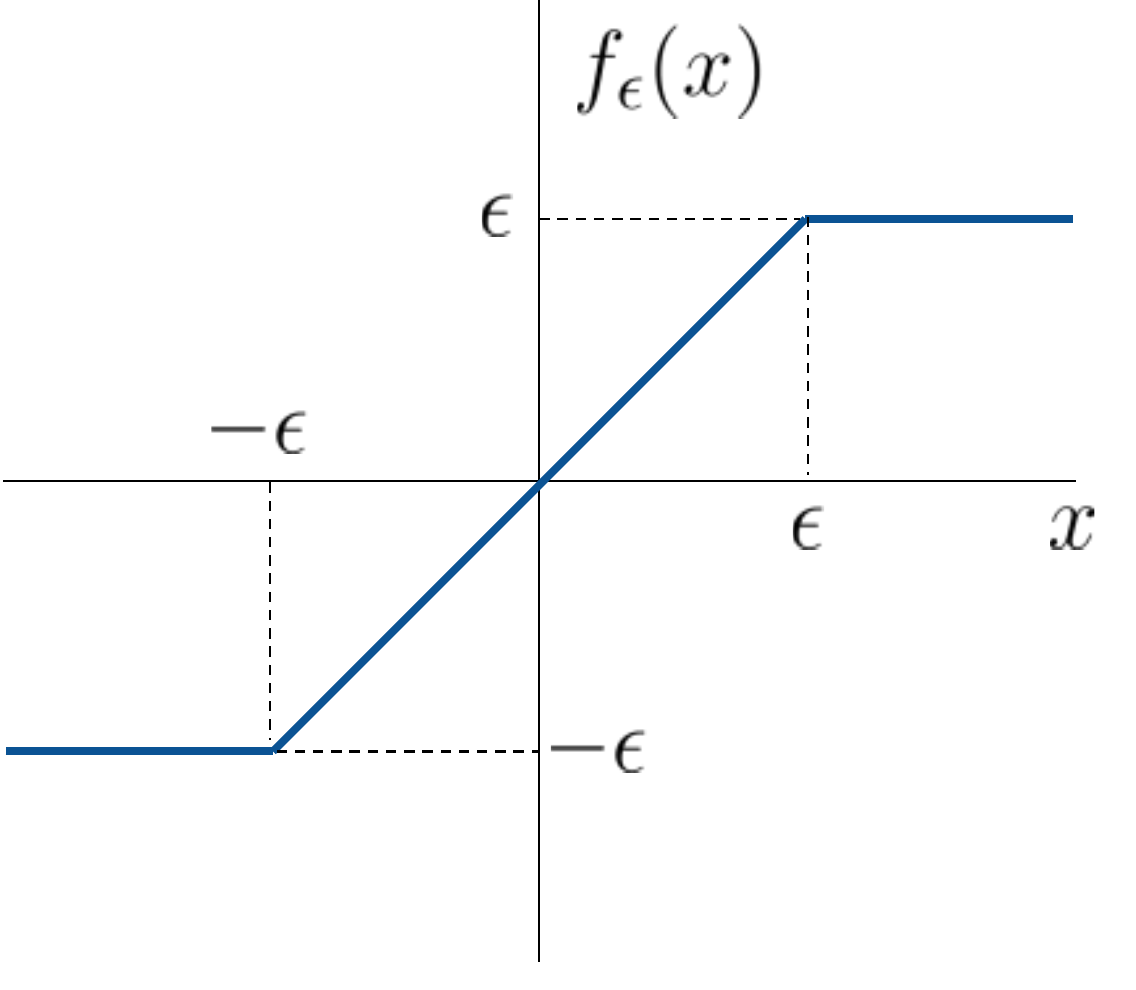}
		\caption{}
		\label{fig:f_eps}
	\end{subfigure}%
	\caption{The double-sided ReLU $g_{\epsilon}()$ and its ``complement'' $f_{\epsilon}()$.}
	\label{fig:fg_func}
\end{figure}

Next, we briefly illustrate how the GLRT defense operates. For a simple setting of binary Gaussian hypothesis testing with equal priors and symmetric means, it is known that the minimax optimal classifier, which achieves the adversarial risk (\ref{eqn:adv_risk}), is a linear classifier of the form $\mathbf{w} = g_{\epsilon} \left( \bm{\mu} \right)$, where $\bm{\mu}$ is the class mean or the ``signal template", $\epsilon$ is the attack budget, and $g_{\epsilon}(\cdot)$ is the ``double-sided ReLU" function, as shown in Fig~\ref{fig:g_eps}. The minimax classifier discards the signal coordinates of low strengths, specifically those whose signs could be flipped when the full attack budget is employed by the adversary. It retains the other coordinates after shrinking them by assuming the worst-case attack has been used, and provides a minimax optimal rule based only on these coordinates. In contrast, the proposed GLRT defense utilizes the signal strength in all the coordinates and applies the double-sided ReLU on a function of the received signal and template. Since GLRT estimates the perturbation, it adapts better when a weaker attack is employed, while minimax schemes are too pessimistic. This is the reason for better robustness-accuracy trade-off of our defense for different attack budgets.

It is worth noting the difference between our approach and classical robust hypothesis testing~\cite{huber,martin,gul}.  In the latter, the focus is typically on resilience to outliers modeled as noise which is independent of the signal corresponding to the true class, whereas our focus is on signal-dependent noise chosen adversarially. Traditionally in hypothesis testing, the actual noise realization is not known and the strategies are noise-agnostic. But it is also of interest to think of noise-aware adversaries, because in machine learning settings, the adversary actually observes the data sample.

\vspace*{0.2cm}
\noindent
{\bf Contributions:}
We summarize our contributions as follows:
\begin{itemize}
	\item The well-known GLRT is proposed as a general approach to defense, in which the desired class and the perturbation are estimated jointly. The GLRT approach applies to any composite hypothesis testing problem~\cite{poor}, unlike minimax strategies optimizing for worst-case attacks, which are difficult to find.
	\item We compare the performance of the GLRT defense to a minimax strategy by considering binary Gaussian hypothesis testing with $\ell_{\infty}$ bounded attacks, for which the minimax strategy has been recently derived \cite{bhagoji}. We demonstrate via an asymptotic analysis and by numerical evaluations that the GLRT approach provides competitive robustness when the attacker employs the full attack budget, while providing better robustness-accuracy trade-offs for weaker attacks. 
	\item We illustrate via examples the application of the GLRT approach to multi-class settings for which minimax strategies are not known. We also provide an intuitively pleasing extension of the binary minimax classifier, which we term as the Pairwise Robust Linear classifier, to benchmark GLRT’s performance. 
	\item We distinguish between noise-agnostic attacks (in which the attacker knows the correct hypothesis but not the noise realization) and noise-aware attacks (in which the attacker knows both the correct hypothesis and the noise realization). For the binary setting, we derive the worst-case attack for the GLRT defense, showing that the same attack is optimal for both noise-aware and noise-agnostic adversaries. For the multi-class setting, we provide a numerical approach for finding the optimal noise-aware attack, and a heuristic noise-agnostic attack which is close to the worst case at high SNR.
\end{itemize}

\noindent
{\bf Notation:}
Throughout the paper, we represent vectors in boldface letters and scalars in regular letters. The norm $|| \cdot ||$ denotes $\ell_2$ norm unless specified otherwise. We denote $\mathcal{N}(\bm{\mu}, \bm{\Sigma})$ as the multivariate Gaussian distribution with mean vector $\bm{\mu}$ and covariance matrix $\bm{\Sigma}$. The symbols $\phi(\cdot)$, $\Phi(\cdot)$, and $Q(\cdot)$ represent the standard (zero-mean, unit variance) univariate Gaussian distribution, its cumulative distribution function (CDF) and the complementary CDF respectively.

\section{Related Work}
\label{sec:related_work}

There is a growing body of research on developing provable robustness guarantees against adversarial attacks. A provably robust defense was developed in~\cite{liang1, liang2}, which employs semidefinite programs and tight relaxations to train neural networks. The idea here is that although it is desirable to find defenses for all possible attacks, computation of worst case error is intractable, hence an upper bound is optimized as a regularizer during training. Another certifiable defense~\cite{wong, wong2} is based on linear programming and optimizing over relaxed convex networks to bound the robustness. Other methods such as in~\cite{hein} obtain guarantees under $\ell_2$ attacks via a regularization functional, for small neural networks. Sparsity is exploited in~\cite{marzi} to provide a theoretical framework that guarantees robustness against $\ell_{\infty}$ attacks on linear classifiers, by introducing a front-end that attenuates the impact of adversary. 

More recently, performance in the presence of adversarial attacks has also been studied through the lens of signal processing, such as in~\cite{jin,li}. Adversarial robustness is investigated in~\cite{jin} by formulating a minimax hypothesis testing problem in situations where the attacker knows the true underlying hypothesis and also when it is unaware of the true hypothesis. \cite{li} studies the robustness of subspace learning problems, such as principle component analysis, where data is modified by an adversary intentionally. There has also been interest in studying adversarial attacks and defense for stochastic bandit algorithms~\cite{liu}.

Among the papers seeking fundamental insights, the problem of finding optimal robust classifiers under binary settings is addressed in~\cite{hassani}, where the class conditional distributions are Gaussian and possess symmetric means. Optimal robust classifiers are derived for binary and ternary classification problems when the perturbations are $\ell_2$ norm-bounded. For the case when perturbations are $\ell_{\infty}$ norm bounded, they restrict attention to the class of linear classifiers and then obtain optimum robust linear classifiers among the restricted class. In general, finding robust optimal classifiers for $\ell_{\infty}$ norm bounded adversarial perturbations is not easily tractable. Analytical results have been shown only for special cases, such as in \cite{bhagoji}, where minimax optimal robust classifiers are characterized in binary classification setting under Gaussian models with symmetric means, same covariance matrices and uniform priors, using ideas from optimal transport theory. They also prove lower bounds on adversarial risk when the classifier is from a certain hypothesis class and compare the performance of robustly trained classifiers for datasets with their bounds. A recent work~\cite{delgosha} proposes a robust classification algorithm under binary Gaussian setting for $\ell_0$ adversaries. Another recent paper~\cite{pydi_20} investigates optimal adversarial risk and classifiers by employing optimal transport theory. Comparable to~\cite{bhagoji}, they characterize optimal adversarial risk for Gaussian models via optimal couplings, and apply similar strategies to find optimal classifiers for univariate uniform and triangular distributions too. 

This work is an extension of the preliminary results reported in our paper~\cite{puranik}, where we introduce the GLRT approach for robustness against adversarial perturbations. In the current paper, we derive the worst-case attack against GLRT defense for binary classification problems, provide performance analysis and further show that the analysis is asymptotically exact. In addition, to characterize GLRT classifier's robustness in multi-class hypothesis testing problems, we provide a procedure to obtain a heuristic based attack that is close to optimal in high SNR regime and illustrate the performance through examples. We also give a method to identify an optimal noise-aware attack for multi-class settings.

\section{GLRT-Based Defense}
\label{sec:glrt_des}
Consider the following standard classification or hypothesis testing problem: 
$$\mathcal{H}_k: \mathbf{X} \sim p_k(\mathbf{x}).$$ 
The presence of an adversary increases the uncertainty about the class-conditional densities, which can be modeled as a composite hypothesis testing problem:
\begin{equation*}
\mathcal{H}_k: \mathbf{X} \sim p_{\theta}(\mathbf{x}), \theta \in \Theta_k,
\end{equation*}
where the size of the uncertainty sets $\Theta_k$ depends on the constraints on the adversary. The GLRT defense consists of joint maximum likelihood estimation of the class and the adversary's parameter:
\begin{equation*}
\hat{k} = \arg\max\limits_{k} \max\limits_{\theta \in \Theta_k} {\it{p}}_{\theta}(\mathbf{x}).
\end{equation*}

\noindent
{\bf Gaussian hypothesis testing:}	
We now apply this framework to Gaussian hypothesis testing with an adversary which can add an $\ell_{\infty}$-bounded perturbation $\mathbf{e}$:
$|| \mathbf{e} ||_{\infty} \leq \epsilon$, where we term $\epsilon$ the ``attack budget'' or ``adversarial budget''.
\begin{equation*}
\mathcal{H}_k: \mathbf{X} = \bm{\mu}_k + \mathbf{e} + \mathbf{N},
\end{equation*}
where $\mathbf{X} \in \mathbb{R}^d$ and $\mathbf{N}\sim \mathcal{N}(\bm{0}, \sigma^2 I_d)$. We assume that the adversary has complete access: it knows the true hypothesis, class mean and could also be aware of the noise realization. 

Conditioned on the hypothesis $k$ and the perturbation $\mathbf{e}$, the negative log likelihood is a standard quadratic expression.  Applying GLRT, we first estimate $\mathbf{e}$ under each hypothesis:
\begin{equation*}
\hat{\mathbf{e}}_k = \arg\min\limits
_{\mathbf{e}: ||\mathbf{e}||_{\infty} \leq \epsilon} ||\mathbf{X} - \bm{\mu}_k - \mathbf{e}||^2,
\end{equation*}
and then plug in to obtain the cost function to be minimized over $k$: 
\begin{equation} \label{Ck_1}
C_k =  ||\mathbf{X} - \bm{\mu}_k - \hat{\mathbf{e}}_k||^2
\end{equation}

This yields illustratively pleasing answers in terms of the function 
$$g_{\epsilon}(x) \triangleq \text{sign}(x) \text{max} \left( 0, |x|-\epsilon \right),$$ 
which we term as the ``double-sided ReLU'' and 
its ``complement,'' $f_{\epsilon}(x) = x - g_{\epsilon}(x)$, shown in Fig.~\ref{fig:fg_func}.  
The estimated perturbation under hypothesis $k$ is obtained as
$\hat{\mathbf{e}}_k = f_{\epsilon}\left( \mathbf{X} - \bm{\mu}_k \right)$, where the non-linearity is applied coordinate-wise.  Substituting into (\ref{Ck_1}),
we obtain 
\begin{equation} \label{eqn:glrt_cost}
C_k = ||g_{\epsilon} \left( \mathbf{X} - \bm{\mu}_k \right) ||^2
\end{equation}
where $g_{\epsilon}(.)$ is applied coordinate-wise. Thus, the GLRT detector 
\begin{equation}
\hat{k} = \arg\min_k C_k
\label{eqn:mc_glrt}
\end{equation}
is a modified version of the standard minimum distance rule where the coordinate-wise differences between the observation and template are passed through the double-sided ReLU.

\vspace*{0.2cm}
\noindent
{\bf Minimax formulation:} An alternative to the GLRT defense, which treats adversarial perturbation as a ``nuisance parameter'' is a game-theoretic formulation. 
Let $\mathcal{H}$ denote the true hypothesis and $\mathcal{\hat{H}}$ be a classifier. The adversary attempts to maximize the probability of error by choosing a suitable perturbation, while the defender tries to choose a classifier such that the expected probability of error is minimized. We consider the perturbations $\mathbf{e}: ||\mathbf{e}||_{\infty} \leq \epsilon$. Thus the optimum adversarial risk is:
\begin{equation}
R^* = \min\limits_{\mathcal{\hat{H}}} \mathbf{\mathbb{E}}\big[\sup\limits_{\mathbf{e} : ||\mathbf{e}||_{\infty} \leq \epsilon } \mathbbm{1}(\mathcal{\hat{H}(\mathbf{X})} \neq \mathcal{H}(\mathbf{X})) \big].
\label{eqn:adv_risk}
\end{equation}
Clearly, this is the best possible approach for defending against worst-case attacks.  Unfortunately, such minimax games are difficult to solve, unlike the more generally applicable GLRT approach. Furthermore, the optimal minimax solution
may be overly conservative, unnecessarily compromising performance against attacks that are weaker than, or different from, the worst-case attack.
In such scenarios, we expect the GLRT approach, which estimates the attack parameters, to provide an advantage.

\section{Binary Gaussian Hypothesis Testing}
\label{sec:analysis}
We now focus on the binary hypothesis testing problem with symmetric means and equal priors for which the minimax rule is known \cite{bhagoji}:
\begin{eqnarray*}
	\mathcal{H}_{0}&:& \mathbf{X} = \bm{\mu} + \mathbf{e} +\mathbf{N} \\
	\mathcal{H}_{1}&:& \mathbf{X} = -\bm{\mu} + \mathbf{e} +\mathbf{N}
\end{eqnarray*}
where $\mathbf{e}$ is chosen by an  $\ell_{\infty}$ bounded adversary, with adversarial budget $\epsilon$, who knows the true hypothesis. 

\vspace*{0.2cm}
\noindent
{\bf Noise-aware v/s noise-agnostic adversaries:}
We consider both noise-aware and noise-agnostic adversarial settings. When the adversary knows the noise realization $\mathbf{N}$, given a classifier $\mathcal{\hat{H}}$ and the true hypothesis $\mathcal{{H}}$, the worst-case adversarial attack causes misclassification whenever possible, depending on the noise realization. The noise-aware formulation of the worst-case attack which is optimal from the adversary's point of view is as below:
\begin{equation}
\mathbf{e}^* = \arg\sup\limits_{\mathbf{e} : ||\mathbf{e}||_{\infty} \leq \epsilon } \mathbbm{1}(\mathcal{\hat{H}}(\mathbf{X}) \neq \mathcal{H}(\mathbf{X})).
\end{equation}

If the adversary does not have access to the noise realization, the optimal attack in the noise-agnostic regime is the maximizer of the class-conditional error, as described below:
\begin{equation}
\mathbf{e}^*_{agn} = \arg\max\limits_{\mathbf{e} : ||\mathbf{e}||_{\infty} \leq \epsilon } \mathbbm{Pr}(\mathcal{\hat{H}}(\mathbf{X}) \neq \mathcal{H}(\mathbf{X})).
\label{eqn:noise_agn_attack}
\end{equation}

\vspace*{0.2cm}
\noindent
{\bf Relation between minimum distance, minimax and GLRT rules:}
We now discuss how the structure of the optimal decision rule without attacks relates to the minimax and GLRT rules.  In the absence of attacks, the optimal rule can be expressed as a minimum distance rule as follows:
\begin{equation}\label{eqn:min_dist_rule_binary} 
|| \mathbf{X} + \bm{\mu}||^2 \gtlt || \mathbf{X} - \bm{\mu}||^2.
\end{equation}
This minimum distance rule can alternatively be expressed as a linear detector which correlates the ``signal template" $\bm{\mu}$ with the observation:
\begin{equation*} \label{linear}
\bm{\mu}^T  \mathbf{X} \gtlt 0.
\end{equation*}
 
As shown in \cite{bhagoji}, the minimax decision rule is also a linear detector of the form
\begin{equation*}
 g_{\epsilon} \left( \bm{\mu} \right)^T  \mathbf{X} \gtlt 0.
\end{equation*}
That is, the minimax rule applies the double-sided ReLU to the ``signal template'' $\bm{\mu}$, and then performs the correlation. Thus, it simply ignores signal coordinates which are small enough such that their signs could be flipped using the worst-case attack budget of $\epsilon$, and shrinks the remaining coordinates to provide an optimal rule {\it assuming that the worst-case attack has been applied.} 

On the other hand, the GLRT rule in the above setting simplifies to a simple modification of the minimum distance rule as following:
\begin{equation} \label{glrt_binary}
C_1 = ||g_{\epsilon} \left( \mathbf{X} + \bm{\mu} \right) ||^2 \gtlt C_0 = ||g_{\epsilon} \left( \mathbf{X} - \bm{\mu} \right) ||^2.
\end{equation}
Comparing (\ref{eqn:min_dist_rule_binary}) and (\ref{glrt_binary}), we see that the GLRT rule applies a coordinate-wise double-sided ReLU to the minimum distance (squared) formulation. Since GLRT applies the double-sided ReLU to the difference between the actual observation and signal templates, we expect that, in contrast to the minimax detector, it should be able to adapt if the attack level is lower than the worst-case attack employing the full budget $\epsilon$. 

One of the possible worst-case attacks for the minimax classifier is: $\mathbf{e} = -\epsilon \cdot\text{sign}(\bm{\mu})$ under $\mathcal{H}_0$ and $\mathbf{e} = \epsilon \cdot \text{sign}(\bm{\mu})$ under $\mathcal{H}_1$. We prove in Sec.~\ref{subsec:worst_attack} that the same attack is indeed the worst-case attack for our GLRT defense under binary classification with Gaussian class-conditionals, for both noise-aware and noise-agnostic adversarial settings.

Under this attack, it is easy to see that the ``defenseless'' minimum distance detector makes errors with probability at least half whenever the attack budget satisfies $\epsilon > ||  \bm{\mu} ||^2/||  \bm{\mu} ||_1$. Thus, the system is less vulnerable (i.e., the adversary needs a large attack budget) when the $\ell_1$ norm of $ \bm{\mu}$  is small relative to the $\ell_2$ norm.  That is, signal sparsity helps in robustness, as has been observed before~\cite{bakiskan, marzi}.

\vspace*{0.2cm}
\noindent
{\bf Application to unequal means:}
The analysis in this paper for the GLRT scheme applies, without loss of generality, to asymmetric means (say $\bm{\mu_0$} and $\bm{\mu_1}$), by shifting of coordinates equivalently, leading to the worst case attack of $\mathbf{e} = -\epsilon \cdot\text{sign}(\bm{\mu_0} - \bm{\mu_1})$ under $\mathcal{H}_0$.
We note that the minimax classifier also applies in a setting with generic means $\bm{\mu}_0$ and $\bm{\mu}_1$. By change of coordinates, we can arrive at a symmetric mean setting, where if the attacker employs $\mathbf{e} = -\epsilon \cdot\text{sign}(\bm{\mu_0} - \bm{\mu_1})$ under $\mathcal{H}_0$ as the worst-case attack against a linear classifier with $\mathbf{w} = g_{\epsilon}(\frac{\bm{\mu_0} - \bm{\mu_1}}{2})$, and from the defender's point of view, fixing the classifier to the minimax scheme is still optimal given such an attack. In general, the classifier takes the form:
\begin{equation}
\label{eqn:minimax_rule_generic} 
g_{\epsilon}\Big(\frac{\bm{\mu_0} - \bm{\mu_1}}{2}\Big)^T \Big(\bm{X} - \frac{\bm{\mu_0} + \bm{\mu_1}}{2}\Big) \gtlt 0.
\end{equation}

\vspace*{-0.2cm}
\subsection{Coordinate-wise analysis}
\label{sec:analysis_glrt}
Since the GLRT rule is nonlinear, its performance is more difficult to characterize than that of a linear detector. However, we are able to provide insight via a central limit
theorem (CLT) based approximation (which holds for large number of dimensions $d$). We focus on noise-agnostic adversaries for the analysis.
By the symmetry of the observation model and the resulting symmetry induced on the attack model, we may condition on $\mathcal{{H}}_0$ and the
corresponding attack $\mathbf{e} = -\epsilon \cdot\text{sign}(\bm{\mu})$, and consider $\mathbf{X} = \bm{\mu} - \epsilon \text{sign}(\bm{\mu}) + \mathbf{N}$. The costs are
\begin{eqnarray*}
	C_0 &=& \sum\limits_{i = 1}^d(g_{\epsilon}(-\epsilon \text{sign}(\bm{\mu}[i]) + \mathbf{N}[i]))^2\\ 
	C_1 &=& \sum\limits_{i = 1}^d(g_{\epsilon}(2\bm{\mu}[i]-\epsilon \text{sign}(\bm{\mu}[i]) + \mathbf{N}[i]))^2.
\end{eqnarray*}
and the error probability of interest is $P_e = P_{e|0} = P[C = C_1 - C_0 < 0|\mathcal{{H}}_0]$.

We now perform a coordinate-wise analysis of the cost difference $C[i] = C_1[i] - C_0[i]$, where $C_k[i]$ indicates the contribution in cost $C_k$ from coordinate $i$. Let the mean and variance of $C[i]$ be denoted by $m_{C[i]}$ and $\rho_{C[i]}^2$ respectively. Applying CLT on the sum across coordinates, the error probability can be estimated as:
\begin{equation}
P_e = P_{e|0} = P\big(\sum_{i = 1}^{d}C[i] < 0\big) \approx Q \left( \frac{\sum_{i=1}^d m_{C[i]}}{\sqrt{\sum_{i=1}^d \rho_{C[i]}^2}} \right).
\label{eqn:clt}
\end{equation}
The error probability analysis can be made exact in the limit as $d \rightarrow \infty$, if the Lindeberg's condition is satisfied for CLT to hold for independent, but not necessarily identically distributed random variables. We show in Sec.~\ref{sec:lindeberg} that Lindeberg's condition is indeed satisfied in our setting. 

\vspace*{0.2cm}
\noindent
{\bf Asymptotic equivalence with minimax classifier:}
Consider a particular coordinate $i$, set $C = C[i]$, and let
$\bm{\mu} [i] = \mu$.  Assume $\mu > 0$ without loss of generality: we simply replace $\mu$ by $| \mu |$ after performing our analysis, since 
the analysis is entirely analogous for $\mu < 0$, given the symmetry of the noise and the attack. 
We can numerically compute the mean and variance of the cost difference for the coordinate, $C = \left( g_{\epsilon}(2\mu + N - \epsilon) \right)^2  - \left( g_{\epsilon}(N - \epsilon)\right)^2$, but the following lower bound yields insight: 
\begin{equation}
C  \geq Y \triangleq \mathbbm{1}_{\{N \geq -t \}} (t + N )^2 - N^2,
\label{eqn:y_bound}
\end{equation}
where $t = 2( \mu - \epsilon )$. Note that $t > 0$ ($| \mu | > \epsilon$) corresponds to coordinates that
the minimax detector would retain. The high-SNR ($t/\sigma$ large) behavior is interesting.
For $t > 0$, we can show that $Y \approx t^2 + 2Nt$; these coordinates exhibit behavior similar to the minimax detector.
On the other hand, for $t < 0$, $Y \approx - N^2$; these coordinates, which would have been deleted by the minimax detector, contribute noise in favor of
the incorrect hypothesis (this becomes negligible at high SNR).  These observations indicate that, at high SNR, the performance of the GLRT detector approaches that of the minimax detector under worst-case attack.

\begin{figure}[t!]
	\centering
	\includegraphics[width=\columnwidth]{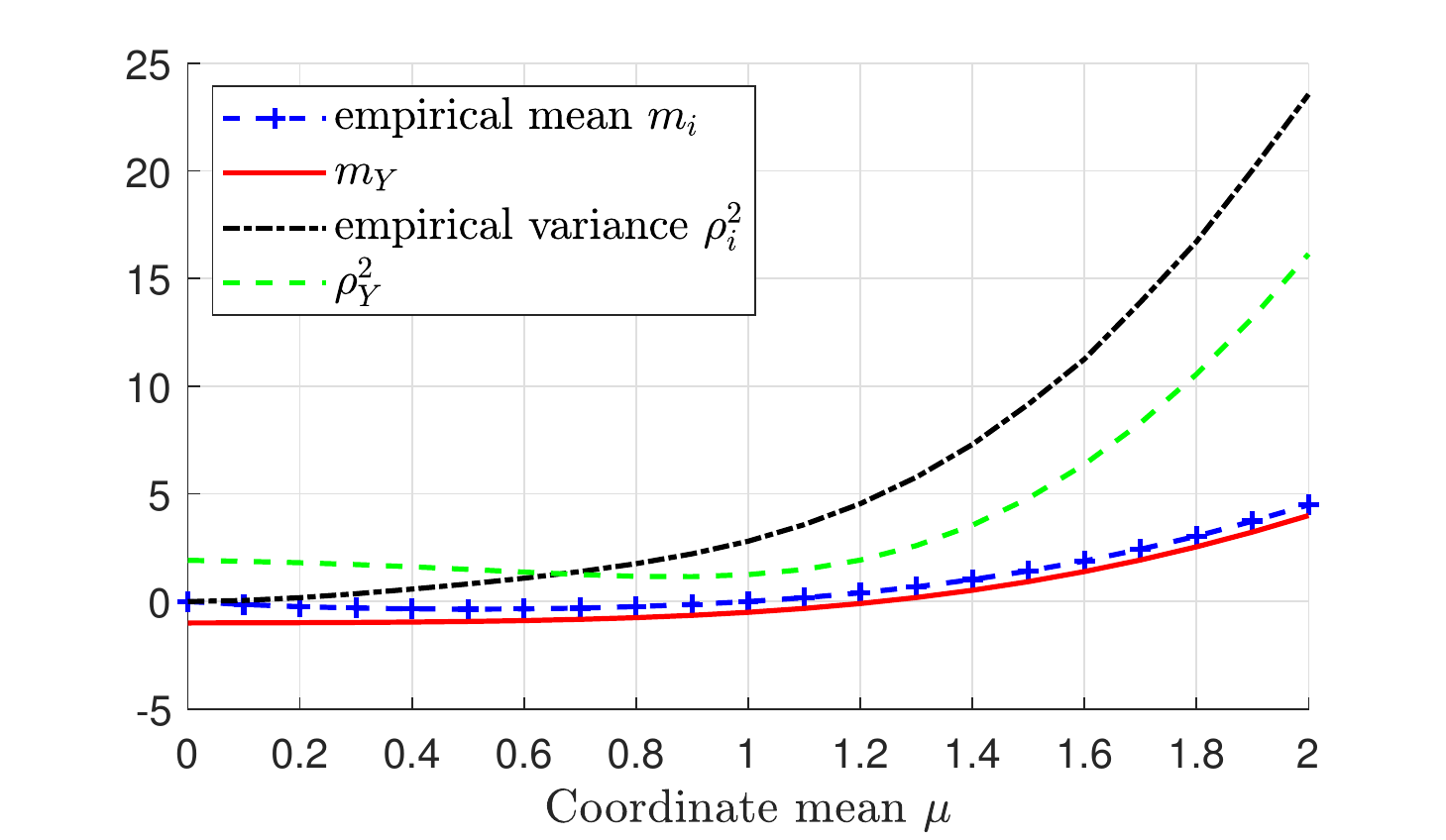}
	\caption{Comparison of empirical mean and variance of $C[i]$ with the mean and variance of  lower bounding variable $Y_i$.}
	\label{fig:coord_mean_var}
\end{figure}

Without loss of generality, let us redefine $t = 2(|\mu| - \epsilon)$. The mean and variance of $Y$, irrespective of $\text{sign}(\mu)$, can be computed in closed form as follows:
\begin{eqnarray}
m_Y &=&  Q\Big(\frac{-t}{\sigma}\Big) (t^2 + \sigma^2) - \sigma^2 + \sigma t \phi\Big(\frac{t}{\sigma}\Big)\label{eqn:mean}\\
\rho^2_Y&=& 3\sigma^4 + Q\Big(\frac{-t}{\sigma}\Big) (t^4 + 4t^2\sigma^2 - 3\sigma^4)\nonumber\\
&& + \sigma t \phi(t/\sigma)(t^2 + 3\sigma^2) - m_Y^2.\label{eqn:var}
\end{eqnarray}
Figure~\ref{fig:coord_mean_var} shows the empirical mean and empirical variance of $C[i]$, i.e., $m_i$ and $\rho^2_i$,  in comparison with $m_Y$ and $\rho^2_{Y}$ obtained through (\ref{eqn:mean}) and (\ref{eqn:var}). Here, the adversarial budget is set to $\epsilon = 1$ and noise variance $\sigma^2 = 1$. 

\vspace*{0.2cm}
\noindent
{\bf GLRT under low noise limit:}
The error probability in (\ref{eqn:clt}) can also be bounded by applying CLT on the lower bounding terms $Y_i \leq C[i]$ as follows:
\begin{eqnarray}
P\big(\sum_{i = 1}^{d}C[i] < 0\big) &\leq& P\big(\sum_{i = 1}^{d}Y_i < 0\big) \nonumber\\ 
&\approx& Q \left( \frac{\sum_{i=1}^d m_{Y_i}}{\sqrt{\sum_{i=1}^d \rho^2_{Y_i}}} \right). 
\label{eqn:clt_low_bd_var}
\end{eqnarray}
Bounding the probability of error in this fashion helps in yielding the following insight. Under low noise limit ($\sigma^2 \rightarrow 0$), the variance $\rho^2_{Y_i} = 0, \forall i$; and the mean is given by $m_{Y_i} = t^2$, if $|\bm{\mu}[i]| > \epsilon$, otherwise it is zero. Thus as long as  $\exists i$ such that $|\bm{\mu}[i]| > \epsilon$, or equivalently $\epsilon < ||\bm{\mu}||_{\infty}$, we have $P_e \to 0$. Interestingly, for the error of the naive minimum distance detector to approach zero under low noise limit, $\epsilon < ||\bm{\mu}||^2/||\bm{\mu}||_1$ should hold, which is a more stringent condition than that required by the GLRT detector. 

Also note that since each of the means and variances are $\mathcal{O}(1)$ terms, we have
$P_e \leq k_1 e^{-k_2 d}$, where $k_1$, $k_2$ are positive constants, irrespective of the SNR requirements. 

\subsection{Asymptotic exactness through Lindeberg's condition}
\label{sec:lindeberg}
The random variables $Y_k$, for $1\leq k \leq d$, are independent, but not identically distributed. For brevity, let the mean and variance of $Y_k$ be denoted by $m_k$ and $\rho^2_k$ respectively. The sum of variances of all the $d$ random variables is given by $s_d^2 = \sum_{k = 1}^{d} \rho_k^2$. A sufficient condition for the central limit theorem (CLT) to hold in the case of independent but not necessarily identically distributed random variables is the Lindeberg's condition.
\begin{proposition}
If the independent, non-identically distributed, random variables $Y_k, k\in[d]$, $\forall\delta>0$ satisfy the following, then the central limit theorem holds.
	\begin{equation}
	\lim\limits_{d \rightarrow \infty} \frac{1}{s_d^2} \sum_{k = 1}^{d} \mathbb{E}\big[ (Y_k - m_k)^2 \mathbbm{1}_{\{|Y_k - m_k| \geq \delta s_d\}}\big] = 0.
	\label{eqn:lc}
	\end{equation}
\end{proposition}
The proof of this proposition is deferred to the Appendix~\ref{sec:lindeberg_proof}. It can further be shown that the Lindeberg's condition is also satisfied by the sum of per coordinate cost differences $C[k]$. Since showing this is analogous, we do not give the detailed case-by-case calculation, but only provide a sketch in Appendix~\ref{sec:lindeberg_proof}. Thus, the approximate equalities in (\ref{eqn:clt}) and (\ref{eqn:clt_low_bd_var}) are indeed exact.

\subsection{Worst-case attack for GLRT defense}
\label{subsec:worst_attack}
In this section, we find the optimal attack from the adversary's point of view, also termed the worst-case attack, given a classifier. Firstly, we note that the worst-case attack for the GLRT defense is not unique. In the following proposition, we show that an attack that is oblivious to the noise realization is also a worst-case attack in the noise-aware setting for binary hypothesis testing under the GLRT classifier.
\begin{proposition}
	\label{prop:glrt_worst_attack}
	A worst-case attack for the GLRT defense in a binary Gaussian classification problem with class means $\bm{\mu}_0$ and $\bm{\mu}_1$, under both noise-aware and noise-agnostic adversarial settings, is given by
	\begin{equation}
	\mathbf{e}^* = - \epsilon \cdot \text{sign}(\bm{\mu}_0 - \bm{\mu}_1), \text{  under $\mathcal{H}_0$}
	\end{equation} 
	\begin{equation}
	\mathbf{e}^* = - \epsilon \cdot \text{sign}(\bm{\mu}_1 - \bm{\mu}_0), \text{  under $\mathcal{H}_1$}.
	\end{equation}
\end{proposition}
\begin{IEEEproof}
	Without loss of generality, let us first consider the symmetric mean case. Following the notation in Sec.~\ref{sec:analysis_glrt}, we first show that for all coordinates where $\mu \geq 0$, the per coordinate cost difference under $\mathcal{H}_0$, given by $C[i]$, is  non-decreasing in $\mathbf{e}[i]$, and where $\mu < 0$, $C[i]$ is decreasing in $\mathbf{e}[i]$. The proof is deferred to Appendix~\ref{sec:app_cost_mono}. Let $\mathbf{e}_1$ and $\mathbf{e}_2$ be two attacks and $\mathbf{N}$ a noise realization. Denoting $\sum_{i}C[i] = \underline{C}$ for brevity, and assuming that $\mu > 0$ for all the coordinates, 
	it follows from the monotonicity of per-coordinate cost difference, that for any fixed $\mathbf{N}$, if $\mathbf{e}_1 \succcurlyeq \mathbf{e}_2$, then $\underline{C}(\mathbf{e}_1, \mathbf{N}) \geq \underline{C}(\mathbf{e}_2, \mathbf{N})$. Let $\mathbf{e}_2 = -\epsilon \cdot \underline{\mathbf{1}}$. For any other attack $\mathbf{e}_1$, and $\forall \mathbf{N}$,
	\begin{equation}
	\underline{C}(\mathbf{e}_1, \mathbf{N}) \geq \underline{C}(-\epsilon\cdot\underline{\mathbf{1}}, \mathbf{N}).
	\end{equation}
	Relaxing the assumption on $\mu$ and utilizing the result that the per coordinate cost difference for indices where $\mu < 0$  is decreasing in corresponding $\mathbf{e}[i]$, it follows that
	\begin{equation}
	\underline{C}(\mathbf{e}, \mathbf{N}) \geq \underline{C}(-\epsilon\cdot \text{sign}(\bm{\mu}), \mathbf{N}) \label{eqn:worst_attack_binary}
	\end{equation}
	for any $\mathbf{e}$ and $\forall\mathbf{N}$. For binary classification, it suffices that the per-coordinate cost difference is negative to cause misclassification. Under $\mathcal{\hat{H}}_0$, we can see from (\ref{eqn:worst_attack_binary}) that the attack $\mathbf{e} = - \epsilon \cdot \text{sign}(\bm{\mu})$ is sufficient to cause misclassification, whenever possible, for any noise realization. Extending to generic means by shift of coordinates, the worst case attack under $\mathcal{H}_0$ is thus given by $\mathbf{e}^* = - \epsilon \cdot \text{sign}(\bm{\mu}_0 - \bm{\mu}_1)$. Since the attack does not utilize the noise realization, it is also the best a noise-agnostic adversary can do.
\end{IEEEproof}

\vspace*{0.2cm}
\begin{observation}\label{obs:min_dist_minimax_nn_aware}
The same attack $\mathbf{e}^* = - \epsilon \cdot \text{sign}(\bm{\mu}_0 - \bm{\mu}_1)$ is the worst-case attack in the presence of both noise-agnostic and noise-aware adversaries under binary settings for minimax and also the naive minimum distance based classifier .

For minimum distance classifier, the cost of choosing hypothesis $\mathcal{H}_i$ is $C_i = || \mathbf{X} - \bm{\mu}_i ||^2$. For binary problems, if a noise-aware adversary wants to cause misclassification, it requires to pick a perturbation such that under $\mathcal{H}_0$, costs are such that $C_1 <C_0$, which reduces to finding a perturbation such that given noise $\mathbf{N}$, 
\begin{equation*}
	\min_{\mathbf{e}: ||\mathbf{e}||_{\infty} \leq \epsilon}(\bm{\mu}_0 - \bm{\mu}_1)^{T}(\mathbf{e} + \mathbf{N}) .
\end{equation*}
However, irrespective of noise, the perturbation  $\mathbf{e} = - \epsilon \cdot \text{sign}(\bm{\mu}_0 - \bm{\mu}_1)$ minimizes the above, due to which it is an optimal attack when the minimum distance classifier is used by a defender. This is also optimal for an agnostic adversary as the attack does not require the knowledge of noise. It is also shown in detail in \textit{Observation}~\ref{obs:min_dist_nn}.
	
Similarly, for the minimax classifier, from (\ref{eqn:minimax_rule_generic}) it can be deduced that the adversary attempts to perform 
	\begin{equation*}
	\min_{\mathbf{e}: ||\mathbf{e}||_{\infty} \leq \epsilon} g_{\epsilon}\Big(\frac{\bm{\mu_0} - \bm{\mu_1}}{2}\Big)^T \big(\mathbf{e} + \mathbf{N}\big)
	\end{equation*}
under $\mathcal{H}_0$, leading to $\mathbf{e} = - \epsilon \cdot \text{sign}(g_{\epsilon}\big(\frac{\bm{\mu_0} - \bm{\mu_1}}{2}\big))$, which is equivalent to $\mathbf{e} =  - \epsilon \cdot \text{sign}(\bm{\mu}_0 - \bm{\mu}_1)$. Thus the same attack is optimal for the three classifiers under binary setting.
\end{observation}

\section{Binary Examples and Discussion}
\label{sec:binary_examples}
Focusing on binary classification problems with symmetric means, let a fraction $p$ of the coordinates have means $\mu = a\epsilon$ and a fraction $(1-p)$ have $\mu = b\epsilon$, where $a>1$ and $0\leq b\leq 1$. Let the designed adversarial budget be $\epsilon$ and the actual attack be $\mathbf{e} = \mp\kappa \text{sign}(\bm{\mu})$, where $0\leq \kappa\leq \epsilon$. For the minimax scheme, note that only the fraction $p$ of the coordinates contribute to signal energy. The decision statistic is $(g_{\epsilon}(\bm{\mu}))^T \bm{X}$, from which it follows that the effective signal-to-noise ratio (SNR) is: 
\begin{equation*}
\text{SNR}_{\text{minimax}} = (a - k)^2 dp\Big(\frac{\epsilon}{\sigma}\Big)^2.
\end{equation*}
For the GLRT scheme, the SNR can be obtained directly from (\ref{eqn:clt}) as 
\begin{equation*}
\text{SNR}_{\text{GLRT}} \approxeq d \frac{(p m_a + (1-p)m_b)^2}{p \rho^2_a + (1-p) \rho^2_b},
\end{equation*}
where $m_a$ and $m_b$ are means, $\rho_a^2$ and $\rho_b^2$ are variances of a single component of $C[i]$ contributed by terms with component means $a\epsilon$ and $b\epsilon$ respectively. The probability of error for both the classifiers is given by $Q(\sqrt{\text{SNR}})$. Note that for the GLRT detector, it is only an approximation as convergence is slow at high SNR, and we need to rely on simulations for more accurate error probabilities.

\begin{figure}[h!]
	\centering
	\includegraphics[width=\columnwidth]{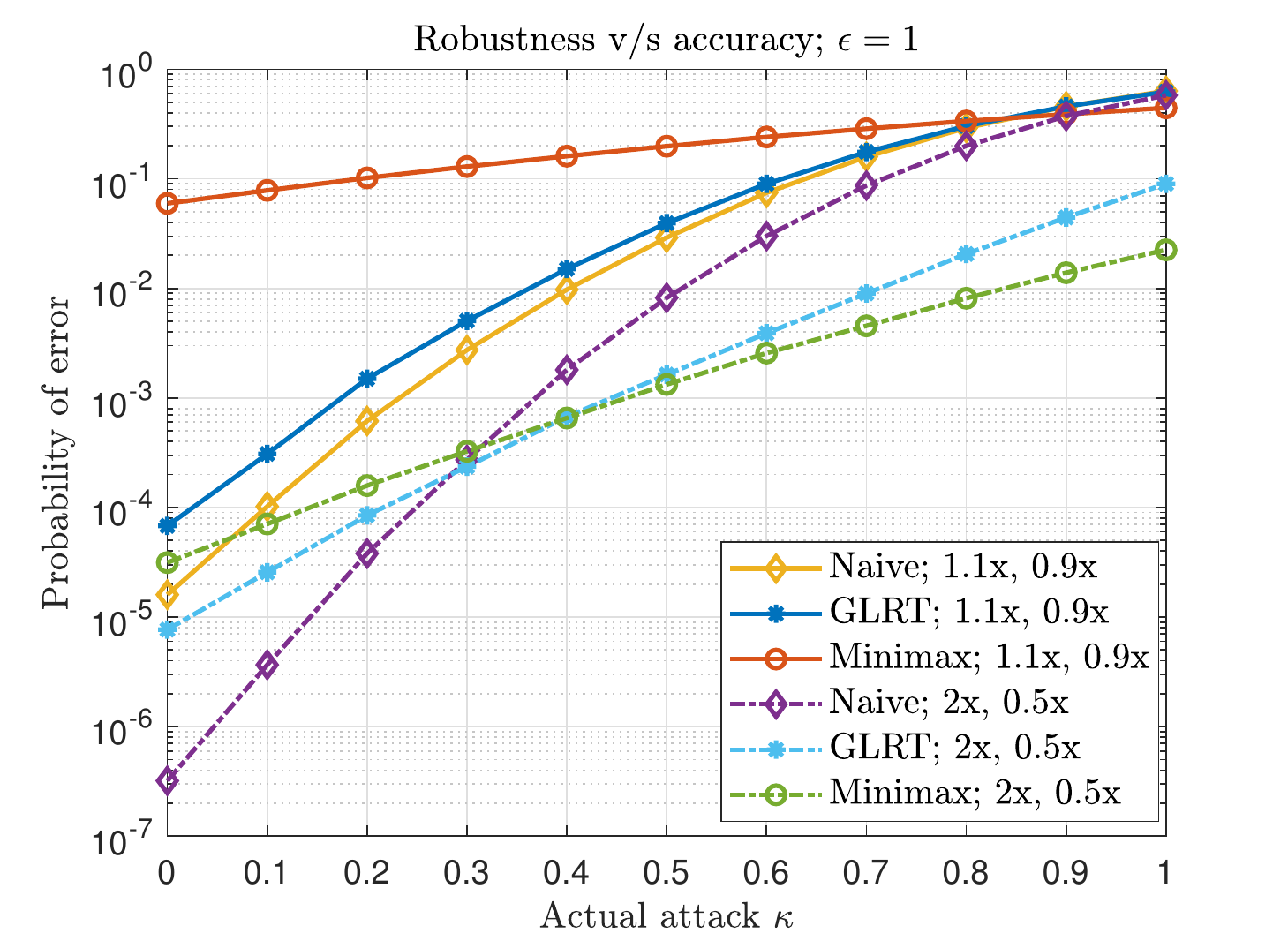}
	\caption{Robustness v/s accuracy trade-off as the actual attack is varied, while the designed adversarial budget is fixed to $\epsilon =1$.}
	\label{fig:rob_acc}
\end{figure}

We consider binary classification problems with symmetric means and uniform priors to draw a comparison with the minimax optimal scheme, and also a naive minimum distance classifier that is optimal under zero attack. The GLRT detector performs better than minimax for weaker attacks, and it has a significant advantage over minimax in settings where the class mean $\bm{\mu}$ has components which are smaller than $\epsilon$, but larger than the actual attack. GLRT utilizes signal energy from these components while for minimax, such components are nulled. Figure~\ref{fig:rob_acc} depicts the performance advantage of GLRT under weaker attacks, for a problem with parameters $\epsilon = 1$, $d = 20$, $p = 0.1$, $a = 1.1$, $b = 0.9$ and noise variance $\sigma^2 = 1$. 

\begin{figure}[h!]
	\centering
	\includegraphics[width=\columnwidth]{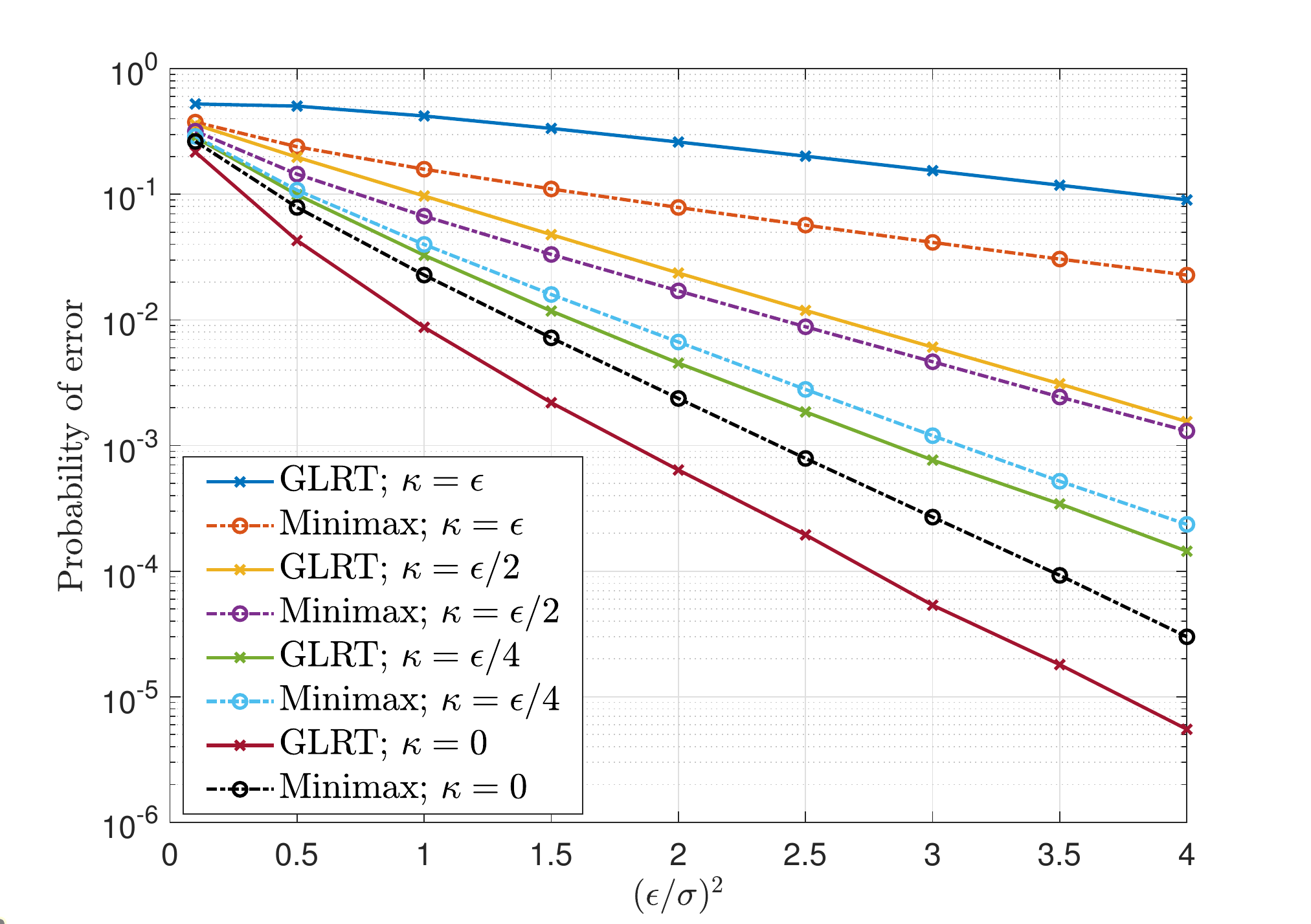}
	\caption{Probability of error as a function of $(\epsilon/\sigma)^2$ for different values $\kappa$ of actual attack (with $\epsilon = 1$, $a = 2$, $b = 0.5$).}
	\label{fig:prob_err}
\end{figure}

The naive minimum distance classifier does poorly under a large attack, specifically in settings where $\bm{\mu}$ has a large number of small components. Under strong attacks, these smaller components contribute to costs in such a way that the wrong class is favored by the naive detector. Consider a problem with parameters $d = 10$, $p = 0.1$, $\epsilon = 1$, $a = 2$, $b = 0.5$ and $\sigma^2 = 0.25$. The comparison of all three detectors under this setting is plotted in Figure~\ref{fig:rob_acc}, which clearly indicates the failure of naive scheme at high attacks, emphasizing the need for a robust detector. Figure~\ref{fig:prob_err} shows the variation of the error probability as a function of $(\epsilon/\sigma)^2$, under four different values of actual attack, for the same problem setting.

\subsection{Speed of convergence}
\begin{figure}[h!]
	\centering
	\includegraphics[width=0.97\columnwidth]{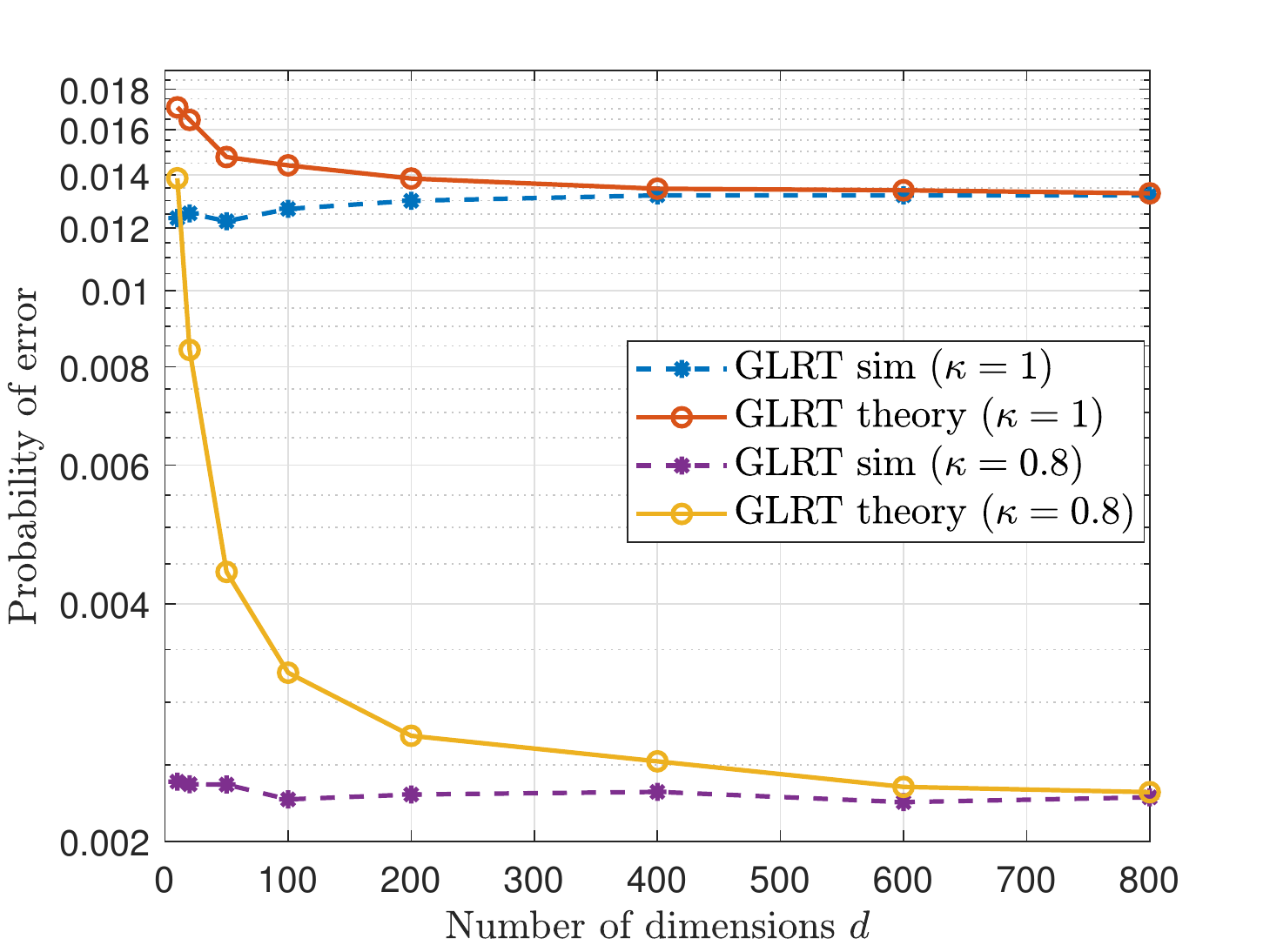}
	\caption{Asymptotic convergence of theoretical error predicted from CLT approximation to the simulation performance of GLRT defense.}
	\label{fig:clt_conv}
\end{figure}

Using CLT to approximate the error probability of the GLRT defense holds only in the limiting case of large $d$. We observe that the distributions of per-coordinate cost differences for each of the coordinates, specifically under low noise, could have narrow asymmetric tails, due to which convergence is slow. We consider a setting with parameters $a = 1.1$, $b = 0.9$, $p = 0.3$, $\epsilon = 1$ and compare the error probabilities as indicated by simulation and those calculated from (\ref{eqn:clt}), where the means and variances of each coordinate of $C[i]$ are computed empirically. We consider two attacks of the form $\mathbf{e} = -\kappa\cdot\text{sign}(\bm{\mu})$, one with the full strength of attack $\kappa = \epsilon = 1$, and another, a weaker attack $\kappa = 0.8$. For both these settings, the true error is fixed to two different values $P_{err_1}\approx Q(\sqrt{5})$ and $P_{err_2} \approx Q(\sqrt{8})$ respectively. Since error is a smooth function of noise variance, the value of $\sigma^2$ for a particular $d$ and the fixed $P_{err}$ is found through grid search. As expected, Fig.~\ref{fig:clt_conv} shows that the theoretical performance approaches that of the simulation as the number of dimensions grows. 

\section{Multi-class hypothesis testing}
The GLRT defense applies to multi-class setting with generic means and priors naturally, as described in (\ref{eqn:mc_glrt}). In order to benchmark the performance of GLRT, we do the following: 
\begin{enumerate}
	\item We first note that deriving a minimax optimal classifier in multi-class setting is a difficult problem, even with the assumption of uniform priors. We consider a heuristic-based extension of the binary minimax classifier, termed the \textit{Pairwise Robust Linear} (PRL) classifier, which we employ to benchmark the performance of GLRT, along with comparing it with the minimum distance classifier.
	
    \item We illustrate that finding an optimal noise-agnostic attack is a difficult problem, and provide a heuristic attack, that is close to the optimal noise-agnostic attack in the high SNR regime, by obtaining a procedure to identify the neighboring class which contributes the most to errors. Using this idea, we arrive at the heuristic based attack in the multi-class setting. 
	
	\item We provide a simple method to identify the optimal noise-aware attack in multi-class problems by extending our knowledge about the optimal noise-aware attack in binary setting. This also gives a lower bound on the classifier's performance.

\end{enumerate}

\vspace*{0.2cm}
\noindent
{\bf Pairwise Robust Linear (PRL) classifier:} Given an $M$-ary classification problem, we can form ${M \choose 2}$ pairs of binary minimax classifiers. The observation is classified as belonging to class $k$ if $k$ is a clear winner in all $M-1$ binary tests $\mathcal{H}_k$ v/s $\mathcal{H}_i$, $i\neq k$, else it is considered an error. We term this as the PRL classifier, since the binary minimax classifier is linear. Note that the PRL classifier need not be minimax optimal. Instead of requiring that a particular class wins against all others, one could also make a decision based on the majority winner among all classes, but for simplicity, we restrict ourselves to requiring a clear winner against all other hypotheses. 

\vspace*{0.2cm}
\noindent
{\bf Noise-agnostic adversary:}
The optimal noise-agnostic attack is described in (\ref{eqn:noise_agn_attack}). It is difficult to obtain in closed-form for non-binary settings. Given $M$ classes, the class conditional error is upper bounded by the sum of errors of pairwise binary hypothesis tests, and at high SNR, we can assume that there is a single competing class that dominates the error calculations. Thus we are interested in finding this competing class.

\vspace*{-0.3cm}
\subsection{Nearest neighbor class determination}
Given $M$ classes, under $\mathcal{H}_j$, one can think of $M-1$ binary classification problems $\mathcal{H}_j$ v/s $\mathcal{H}_i$, where $i \neq j$, and find which of these binary hypothesis tests yields the worst probability of error. At high SNR, the class conditional error for $M$-ary hypothesis testing depends primarily on the worst of the $M-1$ binary hypothesis tests. We term the competing class which yields this worst probability of error as the \textit{nearest neighbor} class. As a proxy for the true worst-case attack in the multi-class setting, one can use the worst-case attack of the binary hypothesis test against the nearest neighbor (NN) class. Therefore, we want to find the NN class, under every hypothesis.

\begin{observation} \label{obs:min_dist_nn}
	The NN class under hypothesis $\mathcal{H}_j$ for the minimum distance classifier is: 
	\begin{equation}
	\label{eqn:min_dist_heur_agnos}
	\hat{k}(j) = \arg\min_k ||\bm{\mu}_{jk}|| - \epsilon \frac{||\bm{\mu}_{jk}||_1}{ ||\bm{\mu}_{jk}||},
	\end{equation}
	where $\bm{\mu}_{jk} = (\bm{\mu}_j - \bm{\mu}_k)/2$.
\end{observation}

We substantiate the above observation as follows. Let us first consider the minimum distance classifier under binary, symmetric means setting. Under $\mathcal{H}_0$, we have $\bm{X} = \bm{\mu} + \mathbf{e} + \bm{N}$, and the linear classifier of the form
$\mathbf{w}_{clean} = \bm{\mu}$ as discussed earlier. The class conditional error simplifies as follows:
\begin{eqnarray}
P_{e|H_0} &=& P(\bm{\mu}^T \bm{X} < 0)\nonumber\\ 
&=& P(||\bm{\mu}||^2 + \bm{\mu}^T \mathbf{e}+\bm{\mu}^T\bm{N} < 0)\nonumber\\
&=& Q\Big(\frac{\bm{\mu}^T \mathbf{e}/||\bm{\mu}||+||\bm{\mu}||}{\sigma}\Big)\nonumber
\end{eqnarray}
and the worst case noise-agnostic attack is:
\begin{equation*}
\mathbf{e}^*_{agn} = \arg\min_{\mathbf{e}: ||\mathbf{e}||_{\infty}\leq \epsilon } \bm{\mu}^T \mathbf{e}/||\bm{\mu}||+||\bm{\mu}||.
\end{equation*}
Through Holder's inequality, we have $\bm{\mu}^T\mathbf{e} \geq - ||\mathbf{e}||_{\infty}||\bm{\mu}||_1 \geq - \epsilon ||\bm{\mu}||_1$, and equality is achieved when $\mathbf{e} = -\epsilon \cdot \text{sign}(\bm{\mu})$. Thus, the error corresponding to the worst attack is of the form
\begin{equation*}
P_{e|H_0} = Q\Big(\frac{||\bm{\mu}|| - \epsilon ||\bm{\mu}||_1 / ||\bm{\mu}||}{\sigma}\Big).
\end{equation*}
In the case of asymmetric means $\bm{\mu}_0$ and $\bm{\mu}_1$, it follows that the worst-case error is: 
\begin{eqnarray}
P_{e|H_0}
&=& Q\Big(\frac{||\frac{\bm{\mu}_0 - \bm{\mu}_1}{2}  || - \epsilon ||\frac{\bm{\mu}_0 - \bm{\mu}_1}{2}||_1 / ||\frac{\bm{\mu}_0 - \bm{\mu}_1}{2}||}{\sigma}\Big)\nonumber\\
&=& Q\Big(\frac{||\bm{\mu}_{01}|| - \epsilon ||\bm{\mu}_{01}||_1 / ||\bm{\mu}_{01}||}{\sigma}\Big),
\label{eqn:min_dist_err_approx}
\end{eqnarray}
where we denote $(\bm{\mu}_0 - \bm{\mu}_1)/2$ as $\bm{\mu}_{01}$. The optimal noise-agnostic attack in binary setting is hence $	\mathbf{e}^*_{agn} = - \epsilon \cdot \text{sign}(\bm{\mu}_0 - \bm{\mu}_1)$, under $\mathcal{H}_0$. Generalizing to a multi-class setting under hypothesis $\mathcal{H}_j$, under high SNR the error probability would be dominated by the binary hypothesis test between class $j$ and another \textit{closest neighbor} class. A heuristic way of proposing an agnostic attack that is close to the optimal agnostic attack is to simply attack that class which contributes the most error. Using (\ref{eqn:min_dist_err_approx}) and the fact that $Q(.)$ is a monotonically decreasing function, the binary test that contributes the largest error is against the class determined in (\ref{eqn:min_dist_heur_agnos}), which is termed as the nearest neighbor class.

This equation also captures that sparsity improves robustness of the naive classifier. For a system with fixed $\ell_2$ norm of the pairwise separation between means $\bm{\mu}_{jk}$, the class with greater $\ell_1$ norm corresponds to the NN class. 

\begin{observation} \label{obs:glrt_nn}
A procedure to identify the NN class for GLRT under hypothesis $\mathcal{H}_j$ is:
\begin{eqnarray}
\hat{k}(j) &=& \arg\min_k \sum_{i:|\bm{\mu}_{jk}[i]|\geq \epsilon}^{}(|\bm{\mu}_{jk}[i]| - \epsilon)^2\label{eqn:glrt_heur1}
\end{eqnarray}
and the attack under hypothesis $\mathcal{H}_j$, that is close to the optimal noise-agnostic attack is $\mathbf{e}_{agn} = -\epsilon \cdot \text{sign}(\bm{\mu}_{j\hat{k}(j)})$. As SNR increases, $\mathbf{e}_{agn}$ approaches $\mathbf{e}^{*}_{agn}$.
\end{observation}

The above is demonstrated as follows. From (\ref{eqn:clt_low_bd_var}), the class conditional error can be upper bounded by using CLT on the bounding variables $Y[i]$. Observe that in the high-SNR regime, the bound in (\ref{eqn:y_bound}) is close to equality. Thus the true probability of error can be approximated as the error found through CLT on the bounding variables. Suppose that a fraction $p$ of the coordinates are such that $|\bm{\mu}_{0k}[i]|\geq \epsilon$. From (\ref{eqn:y_bound}) and following the notation from Sec.~\ref{sec:analysis_glrt}, the mean and variance of $Y[i]$ can be verified to be the following, based on $t_i = 2(\bm{\mu}_{0k}[i] - \epsilon)$ being positive or negative.
\begin{equation*}
m_{Y_i} =
\left\{
\begin{array}{ll}
t_i^2  & \mbox{if } t > 0\\
-\sigma^2 & \mbox{if } t<0 
\end{array}
\right.
\end{equation*}
\begin{equation*}
\rho_{Y_i}^2 =
\left\{
\begin{array}{ll}
4\sigma^2 t_i^2  & \mbox{if } t > 0\\
2\sigma^4 & \mbox{if } t<0 
\end{array}
\right.
\end{equation*}
The error is estimated as the following under high SNR:
\begin{eqnarray}
P_{e|H_0} &\approx& Q \left( \frac{\sum_{i=1}^d m_{Y_i}}{\sqrt{\sum_{i=1}^d \rho^2_{Y_i}}} \right)\nonumber\\
&=& Q \left( \frac{-(1-p)d \sigma^2 + \sum_{i = 1}^{pd}4(|\bm{\mu}_{0k}[i]| - \epsilon)^2}{\sqrt{2(1-p)d \sigma^4 + 4\sigma^2 \sum_{i = 1}^{pd}4(|\bm{\mu}_{0k}[i]| - \epsilon)^2}} \right)\nonumber\\
&\approx& Q \left( \frac{1}{2\sigma}\sqrt{\sum_{i:|\bm{\mu}_{0k}[i]|\geq \epsilon}4(|\bm{\mu}_{0k}[i]| - \epsilon)^2} \right).\nonumber
\end{eqnarray}
Thus, it follows that we can identify the NN class for GLRT under hypothesis $\mathcal{H}_j$ as in Observation~\ref{obs:glrt_nn}, and further simplify as:
\begin{eqnarray}
\hat{k}(j) &=& \arg\min_k \sum_{i:|\bm{\mu}_{jk}[i]|\geq \epsilon}^{}(|\bm{\mu}_{jk}[i]| - \epsilon)^2 \nonumber\\
&=& \arg\min_k \sum_{i:|\bm{\mu}_{jk}[i]|\geq \epsilon}^{}(\bm{\mu}_{jk}[i])^2 - 2 \epsilon(|\bm{\mu}_{jk}[i]|\label{eqn:heur_sparse})\\
&=&\arg\min_k ||g_{\epsilon}(\bm{\mu}_{jk})||^2 \nonumber
\end{eqnarray}
and the attack is $\mathbf{e}_{agn} = -\epsilon \cdot \text{sign}(\bm{\mu}_{j\hat{k}(j)})$.

It is interesting to note that the same coordinates would have been retained by the PRL classifier under the high SNR regime, leading to the same NN class. Note that the analysis above implicitly assumes that the attack utilizes the entire adversarial budget. If the actual attack is a weaker attack of the form $\mathbf{e} = \mp \kappa\cdot\text{sign}(\bm{\mu})$, where $\kappa < \epsilon$, it follows, analogous to (\ref{eqn:y_bound}), that
\begin{eqnarray}
C[i] &=& (g_{\epsilon}(2\bm{\mu}[i] +\bm{N}[i] - \kappa))^2 - (g_{\epsilon}(\bm{N}[i] - \kappa))^2\nonumber\\
&\geq& \mathbbm{1}_{\{\bm{N}[i] \geq -t_i \}} (t_i + \bm{N}[i] )^2 - (\bm{N}[i])^2 \nonumber\\
&\triangleq& Y_i
\end{eqnarray}
where we redefine $t_i = 2\bm{\mu}[i] - \kappa - \epsilon$. 

\begin{observation}
For attacks that are weaker than the designed budget, the NN class for GLRT is given by
\begin{equation}
\hat{k}(j) = \arg\min_k \sum_{i:2|\bm{\mu}_{jk}[i]|\geq \kappa + \epsilon}^{}(2|\bm{\mu}_{jk}[i]| - \kappa - \epsilon)^2.
\label{eqn:glrt_heur_varying_attack}
\end{equation}
\end{observation}

It is interesting to note that method for determining the NN class derived here implicitly depends on the sparsity of separation between the pairwise means $\bm{\mu}_{jk}$, but measured only over the surviving coordinates as expressed by (\ref{eqn:heur_sparse}). 

\vspace*{0.2cm}
\noindent
{\bf Noise-aware adversary:}
When the true class and the noise realization is known, the attacker aims to employ the worst attack that causes misclassification, when possible. Note that given the noise realization and true label, it is not computationally hard to find the worst-case attack. The adversary checks for feasibility, i.e., whether there exists a perturbation that can cause misclassification. However, rather than compute such an attack numerically, we provide a simple procedure to identify the optimal noise-aware attack in multi-class settings, that builds upon the optimal noise-aware attack in binary settings, identified in {\textit{Observation}} \ref{obs:min_dist_minimax_nn_aware}. 

We will now address the question of finding optimal noise-aware attack for $M-$ary classification problems. Recall that the adversary needs to find a perturbation $\mathbf{e}: ||\mathbf{e}||_{\infty} \leq \epsilon$ such that under true hypothesis $\mathcal{H}_i$, the cost for the classifier under some incorrect hypothesis $\mathcal{H}_j$, (given by (\ref{eqn:glrt_cost}) for the GLRT classifier), is smaller than the cost under the true hypothesis $\mathcal{H}_i$. Since the adversary knows the true class and the noise realization, it can compute and check if by employing any of the $M-1$ binary optimal noise-aware attacks for $\mathcal{H}_i$ v/s $\mathcal{H}_j$, $j \neq i$, if the resulting costs are such that $C_j < C_i$, for some $j$. If there exists such a class $j$, then $\mathbf{e} = - \epsilon \cdot \text{sign}(\bm{\mu}_i - \bm{\mu}_j)$ is sufficient to cause misclassification. If such $j$ is not found, then it implies that none of the incorrect class costs can be made small enough, which means it is not possible to cause misclassification in the multiclass setting. Thus using this procedure for every realization of noise seen, the adversary can behave optimally in the noise-aware multiclass setting. The costs under hypotheses are clear for the minimum distance classifier. Since the PRL is an extension of binary minimax classifier, adversary can attack so as to cause at least one of the binary minimax hypothesis tests to fail. Thus the costs for minimax classifier can be used in this procedure to find an optimal noise-aware attack for PRL classifier.

\subsection{Ternary classification examples}
\begin{figure}[t!]
	\centering
	\includegraphics[width=0.9\columnwidth]{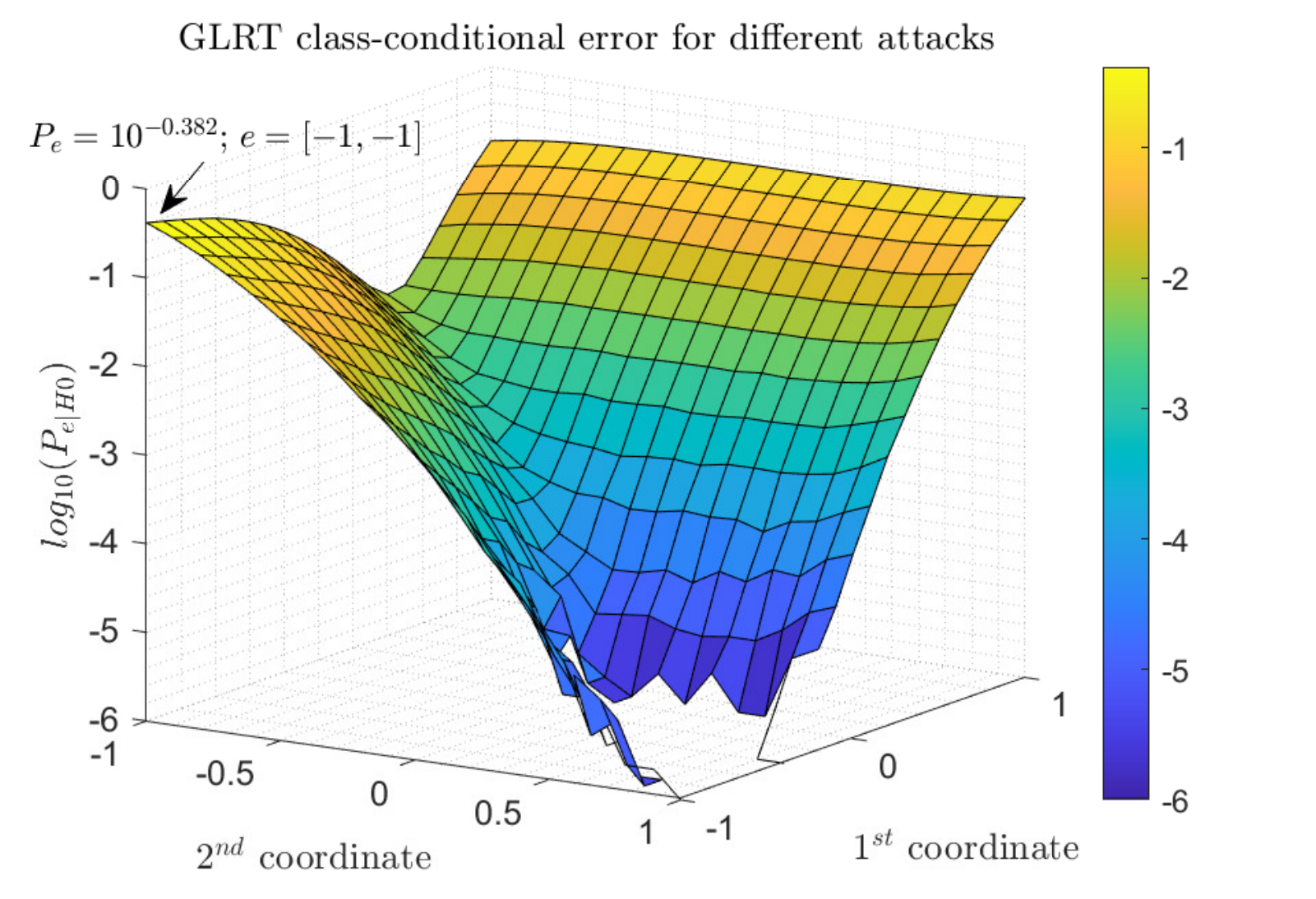}
	\caption{Error surface for the ternary GLRT classifier and its worst case attack.}
	\label{fig:glrt_2d}
\end{figure}

\begin{figure}[t!]
	\centering
	\includegraphics[width=0.9\columnwidth]{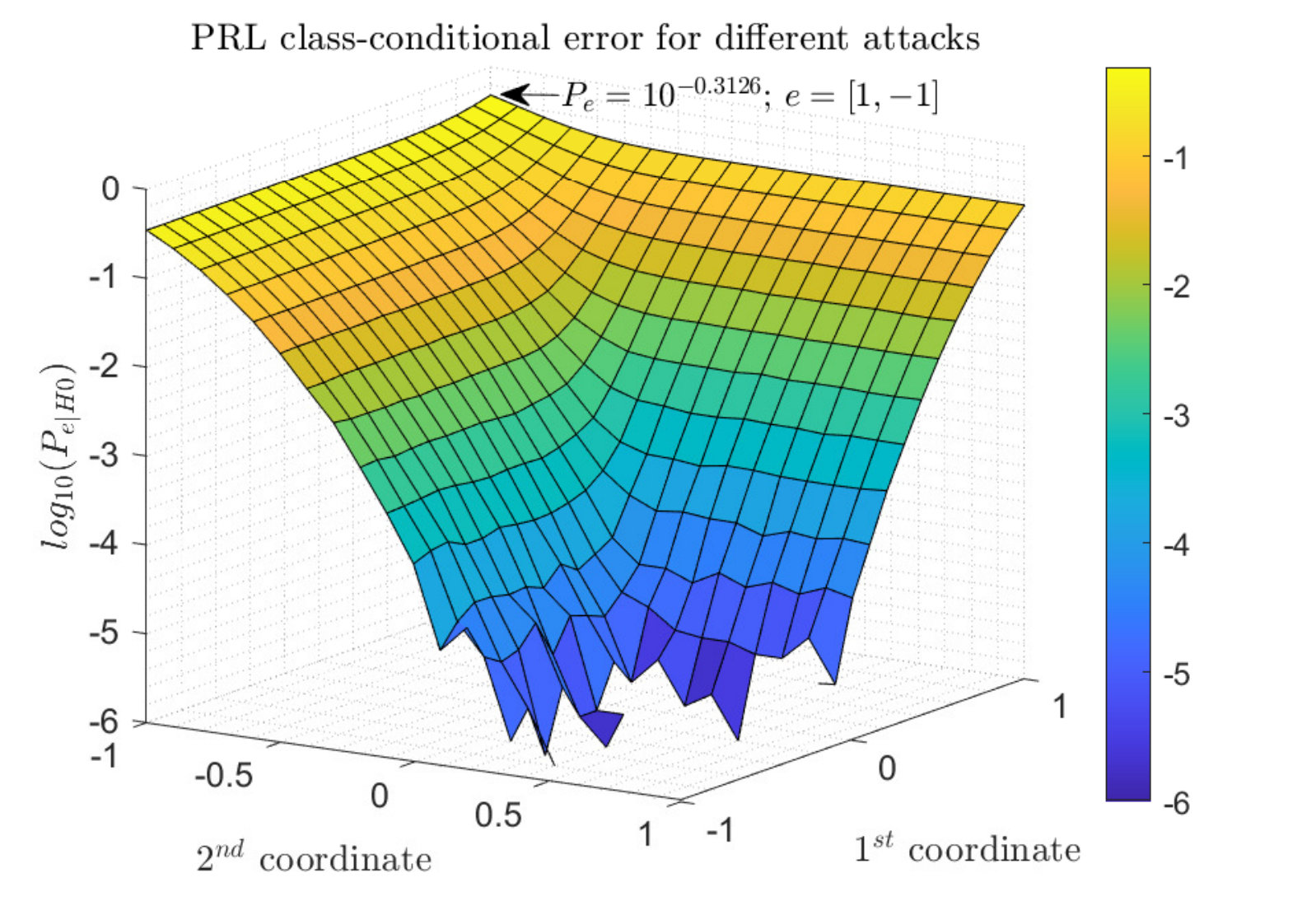}
	\caption{Error surface for the ternary pairwise robust linear classifier and its worst case attack.}
	\label{fig:prl_2d}
\end{figure}

Let us consider a simple two-dimensional ternary classification problem with parameters $\bm{\mu}_0 = [0,0]$, $\bm{\mu}_1 = [2.5,0.25]$, $\bm{\mu}_2 = [-1.75,-2.25]$ and $\sigma^2 = 0.1$, and empirically explore the variation of class-conditional error as a function of the attack, for all valid attacks. Fig.~\ref{fig:glrt_2d} and Fig.~\ref{fig:prl_2d} illustrate the class-conditional error under true class is $\mathcal{H}_0$, for GLRT and PRL classifier respectively. The error surface and the direction of the optimal noise-agnostic attack is different for these classifiers, as indicated in the figures. We also observe that the error surface for GLRT drops faster, for the considered example, in comparison to the error of PRL classifier. The NN class under $\mathcal{H}_0$ as suggested by (\ref{eqn:glrt_heur1}) is $\mathcal{H}_2$, and the corresponding attack $\mathbf{e} = -\epsilon\cdot\text{sign}(\bm{\mu}_0 - \bm{\mu}_2) = [-1,-1]$, which agrees with $\mathbf{e}^{*}_{agn}$ as seen in Fig.~\ref{fig:glrt_2d}. We also checked empirically that the same attack leads to the worst class-conditional error for the PRL classifier, albeit not at the noise level considered in Fig.~\ref{fig:prl_2d}, but at a higher SNR, for the same problem. Though it is simple in a two-dimensional setting to empirically verify that the optimal noise agnostic attack at a particular SNR concurs with the attack suggested by employing NN class calculations, for a large dimensional problem, it is hard to know if the noise variance is low enough for the heuristics to hold, and the optimal noise-agnostic attacks for GLRT, PRL and minimum distance classifiers could be different. 




\begin{figure}[t!]
	\centering
	\includegraphics[width=\columnwidth]{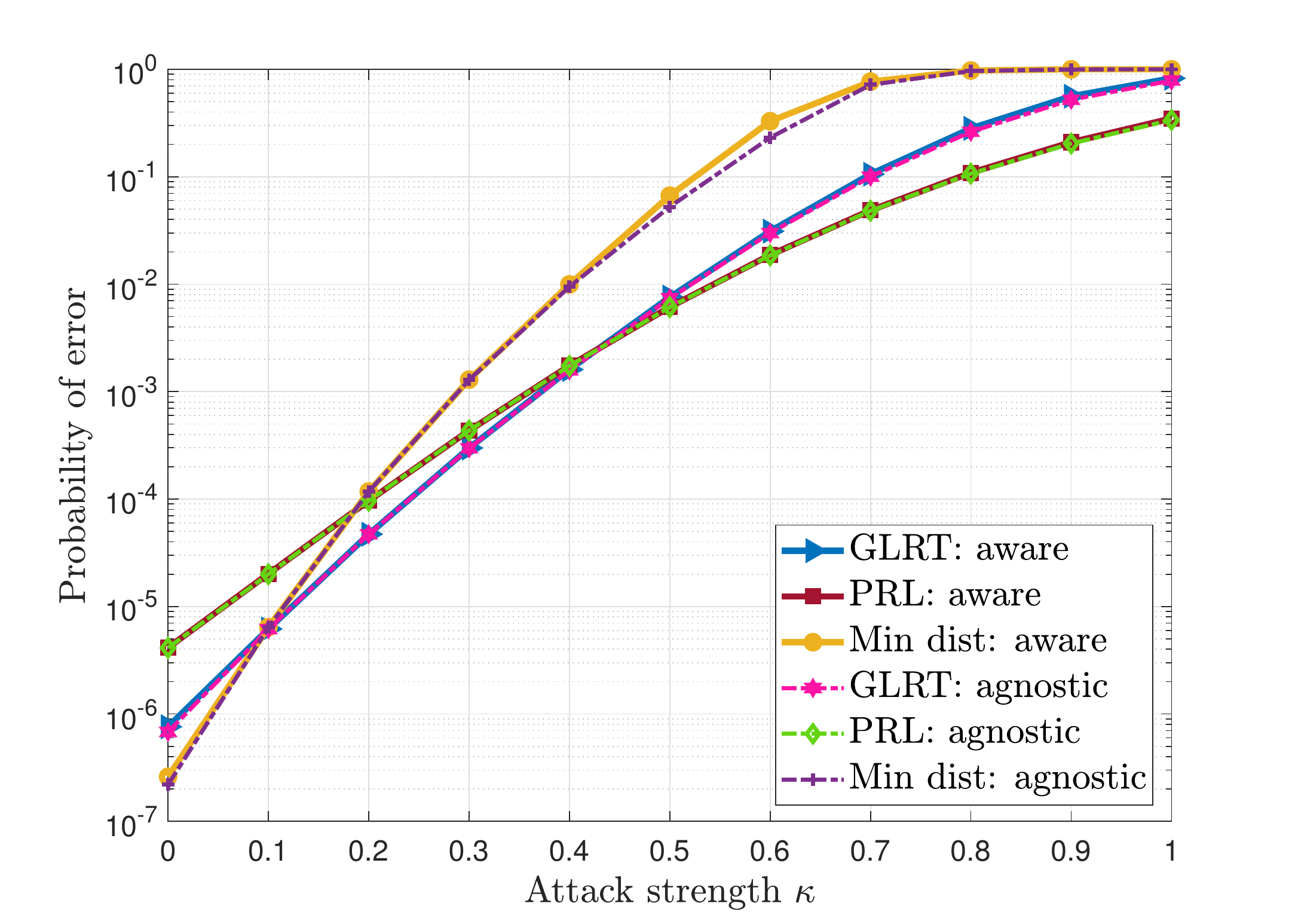}
	\caption{Performance of GLRT, PRL and minimum distance classifiers for the ternary classification problem considered.}
	\label{fig:ternary_aware_agnostic}
\end{figure}

We now consider a ternary setting with equi-probable classes, and parameters $d = 20$, $\epsilon = 1$, noise variance $\sigma^2 = 0.1$, class mean $\bm{\mu}_0$ such that the first $p_0 = 0.15$ fraction of the coordinates are at $0$, and the rest at $1$, i.e., $\bm{\mu}_0 = [0,0,0,1,\ldots, 1]$, $\bm{\mu}_1$ such that the first $p_1 = 0.1$ fraction of the coordinates at $-2.1$ and rest at $0.9$ ($\bm{\mu}_1 = [-2.1,-2.1,0.9,\ldots, 0.9]$), and $\bm{\mu}_2$ such that the first $p_2 = 0.2$ fraction of the coordinates at $-1.8$ and rest at $1.75$ ($\bm{\mu}_2 = [-1.8,-1.8,-1.8,-1.8, 1.75,\ldots, 1.75]$). The strength of the attack is is varied such that $0\leq \kappa \leq \epsilon = 1$. These mean parameters, though seemingly arbitrary, have been chosen such that the pairwise difference between the means possess (i) large number of small components (ii) some components that are smaller the designed budget, but larger than actual attack strength employed. Recall from section~\ref{sec:binary_examples} that settings with these properties showcase simultaneously the quick deterioration of minimum distance classifier as attack strength increases and superiority of GLRT over minimax for weak attacks. This can be observed in Fig.~\ref{fig:ternary_aware_agnostic}, which shows the error probabilities (or error frequencies) of GLRT, PRL and minimum distance classifiers, for both noise-agnostic and noise-aware adversaries. For each of the classifiers, their respective noise-aware optimal attacks are employed to obtain the performance of the classifiers under noise-aware settings. In the case of performance under noise-agnostic adversaries, at each value of the attack strength, the direction of attack is chosen based on the heuristic noise-agnostic attack for each of the classifiers by identifying their respective NN classes (for GLRT, PRL as per (\ref{eqn:glrt_heur_varying_attack}), and for minimum distance classifier as per (\ref{eqn:min_dist_heur_agnos})). For this setting, the performance under aware and agnostic adversaries is not too different, and the gap depends on the separation between class mean parameters and noise variance. The plots for optimal noise-aware attacks also give a lower bound on the performance of the respective classifiers for adversarial hypothesis testing.

\section{Conclusion}
The GLRT approach to adversarially robust hypothesis testing explored in this paper can be generalized to complex models, in contrast to the difficulty of finding optimal minimax classifiers. For the binary model considered here, for which the minimax detector is known, we show that the GLRT detector has the same asymptotic performance as the minimax detector at high SNR for $\ell_{\infty}$ bounded adversarial perturbations at a designated attack level. For attack levels lower than this designated level, the GLRT detector can provide better performance, depending on the specific values of the signal components relative to the attack budget. We derive the worst-case attack for binary settings, analyze the performance, and also show that our analysis is exact as the number of dimensions becomes large. Contrary to minimax, GLRT is a generic multi-class detector that can work with any priors and for multiple classes, as discussed through examples. For multi-class problems, computing an optimal noise-agnostic attack is intractable. However, we obtain a heuristic-based attack that is close to the optimal noise-agnostic attack in the high SNR regime. For noise-aware adversarial settings, a procedure to find optimal noise-aware attacks is provided. 

An interesting direction for future research is to apply the GLRT approach to more complex data and attack models. It is also of interest to explore the minimax formulation in such settings: even if it is difficult to find the optimal minimax rule, a combination of insights from the minimax and GLRT formulations for simpler models might be useful.

\section*{Acknowledgments}
This work was supported by the Army Research Office under grant W911NF-19-1-0053, and by the National Science Foundation under grant CCF 1909320.

\appendices
\section{Lindeberg's Condition for convergence}
\label{sec:lindeberg_proof}
	Denoting $\bm{\mu}[k] = \mu_k$, recall that the difference of costs under the two classes for coordinate $k$ is given by: 
	\begin{eqnarray}
	C[k] &=&  \left( g_{\epsilon}(2\mu_k + N - \epsilon) \right)^2 - \left( g_{\epsilon}(N - \epsilon)\right)^2 \nonumber \\
	&\geq&  \mathbbm{1}_{\{N \geq -t_k \}} (t_k + N )^2 - N^2 \triangleq Y_k \nonumber,
	\label{eqn:yk}
	\end{eqnarray} 
	where $t_k = 2( |\mu_k| - \epsilon )$. 
	
	From \eqref{eqn:mean} and \eqref{eqn:var} we note that the per-coordinate means and variances are finite constants. Let us define the unfavorable event as $B_k = \{|Y_k - m_k| > \delta s_d\}$. Note that the probability of this event is small.
	By Chernoff bounding, it can be shown, for constants $k_1>0$ and $k_2>0$ that
	\begin{equation}
	P(B_k) \leq k_1e^{-k_2\delta s_d}.
	\label{eqn:pb_chernoff}
	\end{equation}
	In the above, we used the fact that in piecewise intervals, $Y_k$ obeys the distribution of polynomials in a Gaussian random variable, and noted that its moment-generating function exists.
	
	The expectation term in (\ref{eqn:lc}) can be split as follows by conditioning on the event $B_k$, and observing that under $B_k^c$, the expectation is zero. Thus, we have
	\begin{eqnarray}
	&&\mathbb{E}\big[ (Y_k - m_k)^2 \mathbbm{1}_{\{|Y_k - m_k| \geq \delta s_d\}}\big]  \nonumber\\
	&=& \mathbb{E}\big[ (Y_k - m_k)^2|B_k\big] P(B_k) \nonumber \\
	&=& \big(\mathbb{E}\big[ Y_k^2| B_k\big] + m_k^2 - 2m_k \mathbb{E}\big[ Y_k| B_k\big]\big) P(B_k)
	\label{eqn:lind_event}
	\end{eqnarray}
	Consider the computation of $\mathbb{E}\big[ Y_k| B_k\big]$. Further conditioning on the event $A_k = \{N \leq -t_k\}$, it simplifies as
	\begin{align}
	\mathbb{E}\big[ Y_k| B_k\big] &= \mathbb{E}\big[ Y_k| A_k, B_k\big] P(A_k|B_k) \nonumber\\&\quad+ \mathbb{E}\big[ Y_k| A_k^c, B_k\big] P(A_k^c|B_k).\label{eqn:cond_Ak}
	\end{align}
	It can be checked from (\ref{eqn:yk}) the definitions of the events that under $A_k$ and $B_k$, $Y_k = -N^2$, governed by the conditions $N \leq -t_k$ and $N < -\sqrt{\delta s_d - m_k}$ for large $d$. Further, for any $\delta > 0$, $d$ can be chosen to be sufficiently large, so that the stricter condition turns out to be the latter. Thus, we have
	\begin{align}
	&\lim\limits_{d \rightarrow \infty}\mathbb{E}\big[ Y_k| A_k, B_k\big] P(B_k) \nonumber \\
	&=\lim\limits_{d \rightarrow \infty} \mathbb{E} \big[-N^2 | N < -\sqrt{\delta s_d - m_k}\big] P(B_k)\nonumber \\
	&=\lim\limits_{d \rightarrow \infty} \big(\frac{-\sigma^2\alpha\phi(-\alpha/\sigma)-\sigma^2\Phi(-\alpha/\sigma)}{\Phi(-\alpha/\sigma)}\big)P(B_k)\nonumber\\
	&=\lim\limits_{d,\alpha \rightarrow \infty} \big(\frac{k_3\alpha}{R(\alpha/\sigma)}+k_4\big)P(B_k), \label{eqn:lind_tmp1}
	\end{align} 
	where $\alpha=\sqrt{\delta s_d - m_k}$ and $k_3$ and $k_4$ are finite constants. The quantity $R(\alpha) = \frac{1-\Phi(\alpha)}{\phi(\alpha)}$ is called the Mills' ratio. For $\alpha > 0$, it has been shown in \cite{gordon_1941}  that $\lim\limits_{\alpha \rightarrow \infty} \alpha R(\alpha) = 1$. Using this fact and (\ref{eqn:pb_chernoff}), we can write
	\begin{equation}
	\lim\limits_{d \rightarrow \infty}\mathbb{E}\big[ Y_k| A_k, B_k\big] P(A_k|B_k)P(B_k) = 0\label{eqn:lind1}
	\end{equation}
	Under the events $A_k^c$ and $B_k$, we have $Y_k = t_k^2 + 2t_k N$, with condition $N > (\delta s_d +m_k -t_k^2)/2t_k$. Thus, we have the following equations:
	\begin{eqnarray}
	&&\lim\limits_{d \rightarrow \infty}\mathbb{E}\big[ Y_k| A_k^c, B_k\big]P(B_k)\\
	&=&\lim\limits_{d \rightarrow \infty} \mathbb{E} \Big[t_k^2 + 2t_k N | N > \frac{\delta s_d +m_k -t_k^2}{2t_k}\Big]P(B_k) \nonumber \\
	&=& \lim\limits_{d,\alpha \rightarrow \infty} \big(k_5+\frac{k_6}{R(\alpha/\sigma)}\big)P(B_k)\nonumber\\
	&=& 0, \label{eqn:lind_tmp2}
	\end{eqnarray} 
	where $\alpha = \frac{\delta s_d +m_k -t_k^2}{2t_k}$ and $k_5$ and $k_6$ are finite constants and we again used the limiting value of the Mills' ratio and the exponential bound on $P(B_k)$. Therefore, we have,
	\begin{equation}
	\lim\limits_{d \rightarrow \infty}\mathbb{E}\big[ Y_k| A_k^c, B_k\big] P(A_k^c|B_k)P(B_k) = 0\label{eqn:lind2}
	\end{equation}
	Similarly, it can be seen that $\lim\limits_{d \rightarrow \infty}\mathbb{E}\big[ Y_k^2| A_k, B_k\big] P(B_k) = 0$ as shown below:
	\begin{align}
	&\lim\limits_{d \rightarrow \infty}\mathbb{E}\big[ Y_k^2| A_k, B_k\big] P(B_k) \nonumber \\
	&=\lim\limits_{d \rightarrow \infty} \mathbb{E} \big[N^4 | N < -\sqrt{\delta s_d - m_k}\big] P(B_k)\nonumber \\
	&=\lim\limits_{d,\alpha \rightarrow \infty} \frac{\sigma^2\alpha^3\phi(\frac{-\alpha}{\sigma})+3\sigma^4(\alpha\phi(\frac{-\alpha}{\sigma})+\Phi(-\alpha/\sigma))}{\Phi(\frac{-\alpha}{\sigma})} \cdot P(B_k)\nonumber\\
	&=0, \label{eqn:lind_tmp3}
	\end{align}
	where $\alpha=\sqrt{\delta s_d - m_k}$. Along similar lines, it can be checked that $\lim\limits_{d \rightarrow \infty}\mathbb{E}\big[ Y_k^2| A_k^c, B_k\big]P(B_k) = 0$. Thus from  \eqref{eqn:lind_event}, \eqref{eqn:lind1}, \eqref{eqn:lind2}, and \eqref{eqn:lind_tmp3} the Lindeberg's condition for CLT holds.	

It can further be shown that the Lindeberg's condition is also satisfied by the sum of per coordinate cost differences $C[k]$. Assuming $\mu_k > \epsilon$, the expressions for $C[k]$ are obtained as
\begin{equation}
\small
C[k] =
\left\{
\begin{array}{ll}
(2\mu_k + N -2\epsilon)^2 - (N - 2\epsilon)^2  &N \geq 2\epsilon \\
(2\mu_k + N -2\epsilon)^2  &0 \leq N \leq 2\epsilon \\
(2\mu_k + N -2\epsilon)^2 - N^2  &2\epsilon - 2\mu_k \leq N \leq 0 \\
-N^2  &-2\mu_k \leq N \leq 2\epsilon - 2\mu_k \\
(2\mu_k + N)^2 - N^2  &N \leq -2\mu_k 
\end{array}
\right.
\label{eqn:ck_val}
\end{equation}
The mean and variance of $C[k]$ are finite, as they involve conditional expectations of Gaussian powers. Following through the steps in the previous proof, computing $\mathbb{E}[C[k]|B_k]$ requires conditioning on the events $A_k^i$, $i\in \{1,2\ldots, 5\}$, considered in the branches of (\ref{eqn:ck_val}). These events partition the sample space of $N$. It can be shown that $\lim\limits_{d \rightarrow \infty}\mathbb{E}\big[C[k]| A_k^i, B_k\big]P(B_k) = 0$ and $\lim\limits_{d \rightarrow \infty}\mathbb{E}\big[C[k]^2| A_k^i, B_k\big] P(B_k) = 0$ analogous to (\ref{eqn:lind1}), and the proof follows.

\section{Monotonicity of per-coordinate cost difference}
\label{sec:app_cost_mono}
The per-coordinate cost difference under $\mathcal{H}_0$, restated below, is given by the following:
\begin{equation}
C = C_1 - C_0 = \left( g_{\epsilon}(2\mu + N +e) \right)^2  - \left( g_{\epsilon}(N +e)\right)^2
\label{eqn:per_coord_cost_diff}
\end{equation}
The above expression can take one of the nine possible values based on the relative values of mean, attack and noise, that determine in which region of the double-sided ReLU their arguments lie. We show that for cases that are valid, the derivative of cost difference with respect to attack is non-negative when $\mu \geq 0$. 

\begin{enumerate}
\item Let us first consider the case when parameters are such that the arguments of double-sided ReLU terms in both $C_1$ and $C_0$ lie in the \textit{negative linear region}, i.e., $2\mu + e+N \leq -\epsilon$ and $e+N\leq -\epsilon$. We have,
\begin{equation*}
C = (2\mu + e+N + \epsilon)^2 - (e+N+\epsilon)^2,
\end{equation*}
and $\partial C/{\partial e} = 4\mu$ is non-negative.

\item $C_1$ in negative linear region and $C_0$ in the \textit{null region} ($2\mu + e+N \leq -\epsilon$ and $-\epsilon \leq e+N\leq \epsilon$) : these conditions are contradictory and the value of $C$ defined by these regions is not legitimate. 
\item Note that similar contradictions result when $C_1$ is in the null region and $C_0$ in \textit{positive linear region} ($e + N \geq \epsilon$). 
\item $C_1$ in positive linear and $C_0$ in negative linear region: we have ${\partial C}/{\partial e} = 4(\mu - \epsilon)$. If $\mu \geq \epsilon$, it is clear that the derivative is non-negative. If $\mu < \epsilon$, the conditions are not simultaneously satisfied, resulting in a contradiction.

\item For the other five cases not explicitly shown, it follows from a simple substitution of the conditions on the arguments of the double-sided ReLU and evaluating the expression for $C$ that $\partial C / \partial e \geq 0$. Thus for any $N$, the per coordinate cost difference $C$ is monotonically non-decreasing in $e$.
\end{enumerate}

Following similar steps, when $\mu < 0$, it can be shown that $C$ is monotonically decreasing in $e$. When $C_1$ is in negative linear and $C_0$ in positive linear regions, $\partial C/\partial e = 4\mu + 4\epsilon$, which is negative when $|\mu| > \epsilon$. Otherwise, we note that the inequalities $2\mu +e+N < -\epsilon$ and $e+N>\epsilon$ are not simultaneously satisfied and this case cannot occur. Similar contradictions occur when i) $C_1$ in null and $C_0$ in negative linear region; ii) $C_1$ in positive linear and $C_0$ in null region of the double-sided ReLU. For all other cases, it is easy to verify that $C$ is decreasing in $e$, if $\mu < 0$.

\ifCLASSOPTIONcaptionsoff
  \newpage
\fi



%
%
%

\bibliographystyle{IEEEtran}
\bibliography{IEEEabrv,refs}

%
%
%
%
%




\end{document}